\documentclass[a4paper,fleqn]{cas-dc}

\usepackage{float}
\usepackage{multirow}
\usepackage{booktabs}
\usepackage{graphicx} 
\usepackage{subcaption}
\usepackage{algorithm}
\usepackage{algpseudocode}
\usepackage{amsmath}
\usepackage{microtype}
\usepackage[figuresright]{rotating}
\usepackage[authoryear]{natbib}

\algnewcommand\algorithmicinput{\textbf{Input:}}
\algnewcommand\algorithmicoutput{\textbf{Output:}}
\algnewcommand\INPUT{\item[\algorithmicinput]}
\algnewcommand\OUTPUT{\item[\algorithmicoutput]}

\def\tsc#1{\csdef{#1}{\textsc{\lowercase{#1}}\xspace}}
\tsc{WGM}
\tsc{QE}

\begin{document}
\let\WriteBookmarks\relax
\def\floatpagepagefraction{1}
\def\textpagefraction{.001}
\let\printorcid\relax 

\shorttitle{Generalized Category Discovery in Federated Graph Learning}    

\shortauthors{Zhongzheng Yuan et al.}

\title[mode = title]{Generalized Category Discovery in Federated Graph Learning}

\author[1]{Zhongzheng Yuan}
\ead{genhz@mail.sdu.edu.cn} 
\credit{Conceptualization, Methodology, Software, Data Curation, Investigation, Formal analysis, Writing -- original draft}

\author[1]{Lianshuai Guo}
\ead{goluis.cc@gmail.com} 
\credit{Software, Writing -- review \& editing}

\author[2]{Xunkai Li}
\ead{cs.xunkai.li@gmail.com} 
\credit{Methodology, Visualization, Supervision, Writing -- review \& editing}


\author[1]{Wenyu Wang}
\cormark[1]
\ead{hochi@sdu.edu.cn} 
\credit{Supervision}

\author[1]{Meixia Qu}
\cormark[1]
\ead{mxqu@sdu.edu.cn} 
\credit{Supervision}


\address[1]{Shandong University, School of Airspace Science and Engineering, Weihai 264209, China}
\address[2]{Beijing Institute of Technology, School of Computer Science and Technology, Beijing 100081, China}

\begin{abstract}
Federated Graph Learning (FGL) enables collaborative learning over distributed graph data, yet existing approaches largely rely on a closed-world assumption, limiting their applicability in dynamic environments where novel categories continuously emerge. To bridge this gap, we target the practical scenario of Federated Graph Generalized Category Discovery (FGGCD), aiming to collaboratively discover novel categories across decentralized graph clients while retaining knowledge of known categories.
We observe that FGGCD introduces two fundamental challenges: (1) the Neighborhood Absorption Effect, where structural fragmentation leads to biased neighborhood aggregation, causing novel nodes to be misclassified as known categories; and (2) Global Semantic Inconsistency, where the aforementioned local biases propagate to the server and are amplified by heterogeneous subgraph distributions, hindering cross-client knowledge integration.
To address these issues, we propose GCD-FGL, an FGL framework for GCD that integrates a client-side Topology-Reliable Semantic Alignment and Discovery process to mitigate the neighborhood absorption effect, and a server-side Hierarchical Prototype Alignment strategy to resolve global semantic inconsistency.
Extensive experiments on five real-world graph datasets demonstrate that GCD-FGL consistently outperforms state-of-the-art baselines, achieving an average absolute gain of +4.86 in HRScore.
\end{abstract}




\maketitle

\section{Introduction}
\label{introduction}

Graph Neural Networks (GNNs)~\cite{jiang2019semi} have become a standard approach for modeling relational data, driving significant advancements across diverse domains including social networks~\cite{guo2020deep}, recommender systems~\cite{yu2022graph, he2023simplifying}, and healthcare~\cite{liu2022graphcdr, wang2023gadrp}. In many practical scenarios, however, graph data is naturally partitioned across multiple distributed data owners, a setting widely observed in applications such as financial and recommendation systems. Due to strict privacy regulations, these data silos cannot be directly centralized. Federated Graph Learning (FGL)~\cite{ZZWLWWLWZ22} offers a viable paradigm for collaborative graph model training under such privacy-preserving constraints.

Despite the remarkable success of FGL, existing paradigms largely rely on a closed-world assumption. This assumes that the label space encountered during training is identical to that during inference. However, in many real-world applications, models are deployed in highly dynamic, open-world scenarios where novel categories continually emerge over time. This fundamental gap between closed-world training and open-world realities restricts the adaptability of existing FGL models to unseen categories in the wild.

\begin{figure}
\centering
\includegraphics[width=1\linewidth]{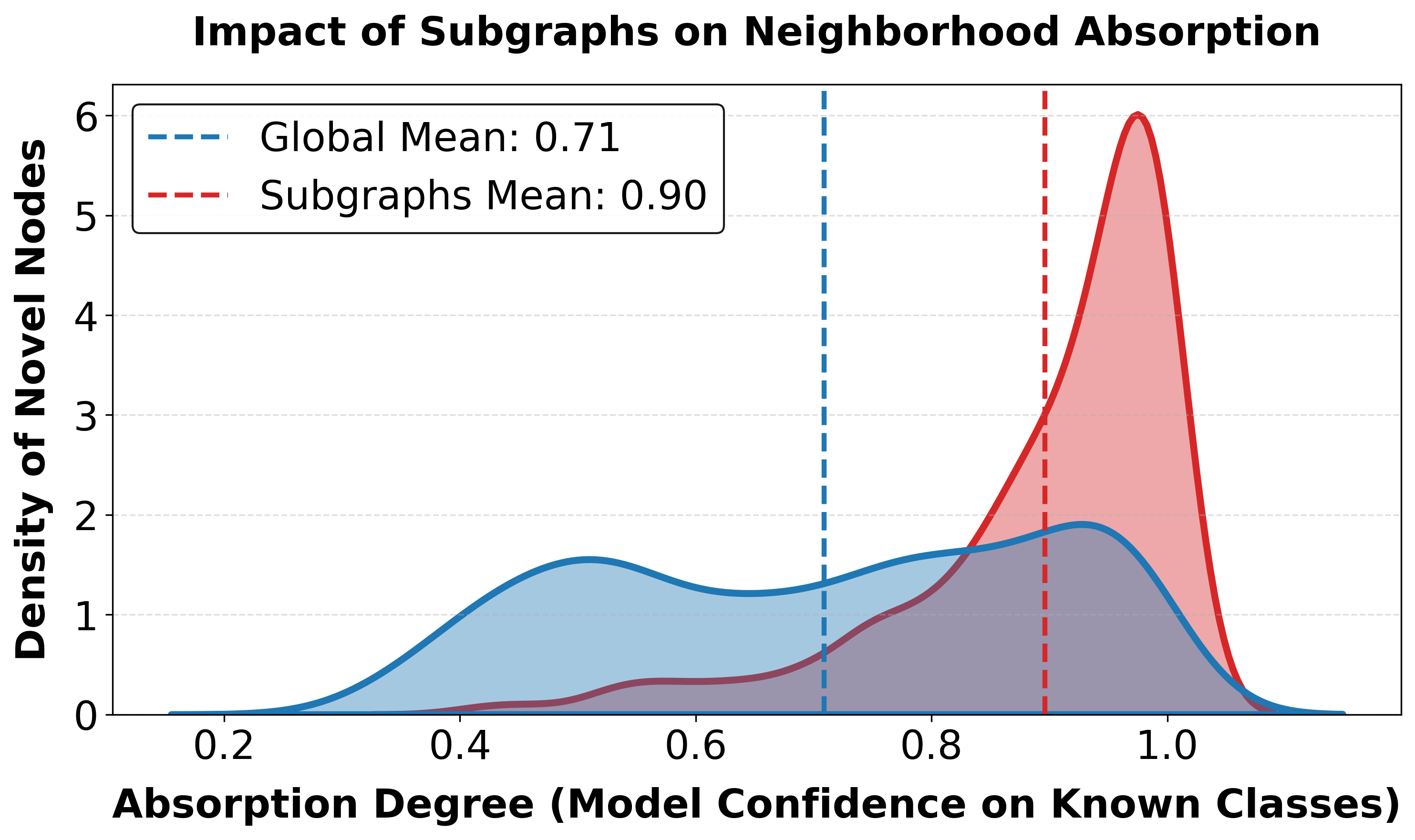}
\caption{The neighborhood absorption effect under a global graph versus isolated subgraphs on Cora. Isolated nodes belonging to novel categories (y-axis) exhibit severe overconfident misclassification towards known categories (x-axis).}
\label{fig:motivation}
\vspace{-0.5cm}
\end{figure}

To take a step towards bridging this gap, Generalized Category Discovery (GCD)~\cite{vaze2022generalized} has been proposed as a critical component of open-world learning. Given an unlabeled dataset drawn from a mixed label space $\mathcal{Y}_{known} \cup \mathcal{Y}_{novel}$, GCD aims to simultaneously classify known categories and discover novel ones, leveraging knowledge from a labeled dataset defined solely on $\mathcal{Y}_{known}$. Unlike Out-of-Distribution (OOD) detection~\cite{hendrycks2017baseline}, which simply rejects unfamiliar samples, or Novel Class Discovery (NCD)~\cite{han2019learning}, which assumes all unlabeled data belongs to $\mathcal{Y}_{novel}$, GCD operates under a much more realistic and challenging open-world setting. It requires the model to precisely distinguish between known and novel categories while concurrently clustering the unknowns into meaningful semantic categories.

Recently, efforts have been made to extend GCD to complex data structures. For instance, Graph Generalized Category Discovery (GGCD)~\cite{deng2025towards} has emerged to tackle category discovery on graph data, assuming a holistic graph, while Federated GCD~\cite{pu2024federated} has been explored for federated learning but only in image data. However, the intersection of GCD and FGL remains largely unexplored. Federated Graph Generalized Category Discovery (FGGCD) aims to collaboratively discover novel categories across distributed graph silos. Yet, simply applying existing GGCD methods to a federated environment fails to work effectively. Specifically, extending GGCD to federated settings faces two primary challenges:

\textbf{Challenge 1: Structural Truncation and the Neighborhood Absorption Effect.} Applying GCD to federated environments introduces a fundamental conflict: GCD requires a holistic data distribution to discover novel clusters, while federated settings physically isolate the graph~\cite{zhang2023federated}. For a novel-category node $v$ ($y_v \in \mathcal{Y}_{novel}$), this structural truncation deprives it of sufficient intra-class support. Consequently, its local observation is frequently dominated by known-category neighbors. During the GNN message-passing process, the learned representation $h_v$ becomes biased toward these known-category signals. This representation collapse, characterized by the \textit{neighborhood absorption} effect, leads to structural semantic dilution: $P(h_v \mid y_v \in \mathcal{Y}_{novel}) \to P(h_v \mid y_v \in \mathcal{Y}_{known})$. 
To empirically validate this, we conduct a motivating study on the Cora dataset (see Figure~\ref{fig:motivation}) comparing representations trained on an intact global graph versus isolated subgraphs. We observe that structural truncation induces an increase in the rate at which novel nodes are absorbed and misclassified into known categories. This absorption degrades the structural semantics essential for GCD exploration, posing an obstacle to reliable novel cluster discovery. To tackle this, we design a client-side Topology-Aware Graph Contrastive Flow module, as further detailed in Section~\ref{method:client}.

\textbf{Challenge 2: Global Semantic Inconsistency Propagated by Local Collapse.} The local representation collapse outlined above propagates to the server, driving and amplifying semantic inconsistency at the global level. Federated graphs inherently exhibit non-IID (non-independent and identically distributed) characteristics~\cite{li2024fedgta, zhu2024fedtad} driven by the homophily assumption, where nodes sharing similar labels concentrate within specific clients. When local representations of novel nodes are distorted by their specific, skewed local known-category neighbors (Challenge 1), the resulting local semantic spaces become largely client-specific. Because novel categories lack explicit supervision and are dynamically explored on local clients, this interplay between local representation offset and graph-induced label skew means that novel clusters formed by different clients often possess divergent semantic meanings. Consequently, accurately matching and aligning these isolated novel clusters at the server side becomes challenging. This challenge is exacerbated in FGGCD scenarios where the total number of global categories is unknown. Without a mechanism to reconcile these misaligned local semantic spaces, the framework is prone to global semantic inconsistency. Within our GCD-FGL framework, we resolve this misalignment through a server-side Hierarchical Prototype Alignment strategy, as detailed in Section~\ref{method:server}.

To address these challenges, we propose GCD-FGL, a federated framework tailored for FGGCD. Specifically, to tackle Challenge 1 on the client side, we introduce a Topology-Reliable Semantic Alignment and Discovery process guided by the Topology Reliability Guidance (TRG) mechanism. This integration effectively calibrates isolated node representations, mitigating the local representation bias and the neighborhood absorption effect caused by structurally truncated subgraphs. To address Challenge 2 on the server side, we couple a Hierarchical Prototype Alignment strategy. This architecture harmonizes the global semantic space, effectively resolving the global semantic inconsistency propagated by local representation collapse across highly heterogeneous label spaces.

The main contributions of this work are as follows:

\textbf{New Observation.} To the best of our knowledge, this is the first work to pioneer and formalize the challenging problem of FGGCD, bridging the gap between GCD and FGL.
\textbf{New Method.} We propose GCD-FGL, a novel framework that integrates a client-side Topology-Reliable Semantic Alignment and Discovery process and a server-side Hierarchical Prototype Alignment strategy, tailored specifically for FGGCD scenarios.
\textbf{SOTA Performance.} We conduct extensive experiments on five real-world graph benchmark datasets. The results show that GCD-FGL consistently outperforms state-of-the-art baselines, achieving an average gain of +4.86 in HRScore.

\section{Preliminaries and Related Work}
\label{sec:related_work_preliminaries}

\subsection{Preliminaries and Problem Formulation}

Table~\ref{tab:notations} summarizes the main notations used in this paper. We consider a global graph $\mathcal{G}$ partitioned into $N$ isolated local subgraphs $\{\mathcal{G}_1, \dots, \mathcal{G}_N\}$ across distributed clients. The global label space $\mathcal{Y} = \mathcal{Y}_{known} \cup \mathcal{Y}_{novel}$ comprises disjoint known and novel categories.

For each client $i$, its local graph $\mathcal{G}_i$ consists of a labeled set $\mathcal{V}_i^L$ (where $y_v \in \mathcal{Y}_{known}$) and an unlabeled set $\mathcal{V}_i^U$ (where $y_v \in \mathcal{Y}_{known} \cup \mathcal{Y}_{novel}$). Specifically, each client contains both labeled known categories and unlabeled samples. Due to data heterogeneity, novel categories are not necessarily shared across clients and may be highly heterogeneous. Under privacy-protected constraints, cross-client edges are unobservable, meaning the union of local edges is a strict subset of global edges ($\bigcup_{i=1}^N \mathcal{E}_i \subset \mathcal{E}$).

The primary goal of FGGCD is to collaboratively learn global parameters $\omega$ that minimize empirical risk across clients without sharing local data. Ultimately, it seeks a unified representation space for both known-category classification and dynamic novel-category discovery, and this balance is quantitatively evaluated by the HRScore. The global objective is formulated as:
\begin{equation}
    \min_{\omega} \sum_{i=1}^N \frac{|\mathcal{V}_i|}{|\mathcal{V}|} \mathcal{L}_{total}(\omega; \mathcal{V}_i^L, \mathcal{G}_i) + \mathcal{L}_{global}(\omega).
    \label{eq:global_obj}
\end{equation}

where $\mathcal{L}_{total}(\omega; \mathcal{V}_i^L, \mathcal{G}_i)$ denotes the local optimization objective evaluated on client $i$ over its local subgraph $\mathcal{G}_i$, guided by the available labeled set $\mathcal{V}_i^L$. To achieve a globally consistent semantic space despite data isolation, $\mathcal{L}_{global}(\omega)$ acts as a conceptual global alignment regularization term. This abstract formulation establishes the theoretical goal of FGGCD; the specific realizations of the local discovery objectives and the global alignment mechanisms will be thoroughly detailed in Section~\ref{method:client} and Section~\ref{method:server}.

\begin{table}[ht]
\centering
\caption{Summary of main notations used in GCD-FGL.}
\label{tab:notations}
\resizebox{\columnwidth}{!}{%
\begin{tabular}{ll}
\toprule
\textbf{Notation} & \textbf{Description} \\
\midrule
\multicolumn{2}{c}{\textit{Graph \& Data Space}} \\
\midrule
$\mathcal{G}$, $\mathcal{G}_i$ & Global graph / Local subgraph on client $i$ \\
$\mathcal{V}$, $\mathcal{E}$ & Global node set / Global edge set \\
$\mathcal{V}_i^L$, $\mathcal{V}_i^U$ & Labeled set / Unlabeled set on client $i$ \\
$\mathcal{N}_i(v)$, $d_v$ & Local neighborhood / Local degree of node $v$ \\
$\mathcal{Y}_{known}$, $\mathcal{Y}_{novel}$& Set of known categories / Set of novel categories \\
\midrule
\multicolumn{2}{c}{\textit{Federated Setting \& Reliability}} \\
\midrule
$N$, $\omega$, $\omega_i$ & Total clients / Global model / Local model \\
$S_{conf}^{(v)}$, $S_{homo}^{(v)}$ & Predictive confidence / Structural smoothness of $v$ \\
$w_v$, $w_i$ & TPR score of node $v$ / Average TPR of client $i$ \\
$\mathbf{c}_i = \{c_{i,k}\}$ & Set of local sample densities on client $i$ \\
$M_v$ & Boolean mask for confident pseudo-labels \\
\midrule
\multicolumn{2}{c}{\textit{Server \& Prototype Management}} \\
\midrule
$\mathcal{P}_{global}$ & Global prototype buffer at the server \\
$P_k^{(t)}$ & Global prototype for category $k$ at round $t$ \\
$\mathcal{P}_{i}^{known}$, $\mathcal{P}_{i}^{novel}$ & \begin{tabular}{@{}l@{}}Local known prototypes / Local novel prototypes \end{tabular} \\
$K_{local}$ & Relaxed local cluster count for prototype extraction \\
$v_{i,k}$ & Joint density-reliability weight \\
$\mathcal{X}_{nov}$ & Global discovery pool of unassigned sub-prototypes \\
$\mathcal{P}_{cand}$, $\mathcal{P}_{hist}$ & Global candidate centers / Historical novel prototypes \\
\midrule
\multicolumn{2}{c}{\textit{Objectives \& Hyperparameters}} \\
\midrule
$\mathcal{L}_{sup}$, $\mathcal{L}_{unsup}$, $\mathcal{L}_{align}$ & \begin{tabular}{@{}l@{}}Supervised Flow / TPR-Guided Unsup. Flow / \\ Dual-flow Semantic Alignment Loss ($\mathcal{L}_{sup} + \mathcal{L}_{unsup}$)\end{tabular} \\
$\mathcal{L}_{gcl}$ & Topology-Aware Graph Contrastive Flow loss \\
$\tau$, $\tau_{sharp}$ & Dynamic temperature / Sharpened temperature \\
$\theta^*$, $\lambda_{hc}$ & Optimal Threshold Cut / Hierarchical penalty \\
$\tau_{base}$, $\tau_{density}$ & Base matching threshold / Density filter threshold \\
$A_{r,c}$ & Boolean assignment matrix for prototype matching \\
$\rho$ & EMA Momentum Update decay factor \\
\bottomrule
\end{tabular}
}
\end{table}

\subsection{Generalized Category Discovery}

The ability to identify both known and novel categories in unlabeled data is crucial for open-world learning. Early paradigms addressing unseen categories primarily focused on OOD detection~\cite{hendrycks2017baseline}, which merely rejects unfamiliar samples without further categorizing them, or NCD~\cite{han2019learning}, which operates under the strict assumption that all unlabeled data belongs to the set of novel categories. To relax strict NCD assumptions, AutoNovel~\cite{han2020autonovel} proposed transferring knowledge from known to novel categories through self-supervised representation learning, but it still struggled to handle realistic scenarios where unlabeled data contain a mixture of known and unknown categories. To address this mixed-data challenge, ORCA~\cite{cao2022open} introduced an open-world semi-supervised learning framework using uncertainty modeling, but its complex multi-stage objective often led to suboptimal feature alignment. Subsequently, the pioneering work GCD~\cite{vaze2022generalized} formally defined the GCD problem and established a strong baseline by using contrastive pre-training followed by non-parametric $K$-means clustering; however, this two-stage pipeline fundamentally decoupled representation learning from cluster assignment. To overcome this disconnect, SimGCD~\cite{wen2023parametric} introduced a parametric framework that jointly optimizes representations and cluster prototypes end-to-end, significantly improving the accuracy of novel-category discovery on image data.

While GCD has been extended to federated settings via works like FedGCD~\cite{pu2024federated}, these advances are strictly confined to independent image data. Existing methods assume either globally intact graphs or independent samples, and these assumptions collapse in FGL. In FGL, geographically distributed yet interconnected data suffers severe structural truncation across client silos. Consequently, discovering novel categories on decentralized graphs with heterogeneous label spaces and severed structural dependencies remains a formidable challenge.

\subsection{Federated Graph Learning}

To collaboratively train GNNs~\cite{kipf2016semi} over distributed subgraphs without centralizing raw data, FedAvg~\cite{mcmahan2017communication} serves as the standard optimization strategy. Specifically, the global aggregation at communication round $t$ is computed as:
\begin{equation}
    \tilde{\omega}^{t} = \sum_{i=1}^N \frac{|\mathcal{V}_i|}{|\mathcal{V}|} \omega_i^{t-1}.
    \label{eq:fedavg}
\end{equation}

where $\tilde{\omega}^{t}$ is the global model, and $\omega_i^{t-1}$ is the local model updated by client $i$ in the previous round. Upon receiving $\tilde{\omega}^{t}$, client $i$ performs local updates via gradient descent:
\begin{equation}
    \omega_i^{t} = \tilde{\omega}^{t} - \eta \nabla f(\tilde{\omega}^{t}, \mathcal{G}_i).
\end{equation}

where $\eta$ is the learning rate and $\nabla f(\cdot)$ denotes the gradient of the Total Loss over the isolated subgraph $\mathcal{G}_i$.

While FedAvg supports isolated training, it performs poorly on non-IID client data and struggles to capture the unique cross-client dependencies of graph-structured data. To address this, FedSage+~\cite{zhang2021subgraph} trained a missing-edge generator to reconstruct cross-subgraph links, mitigating structural truncation, but it focuses mainly on structural completion. Subsequently, FedProto~\cite{tan2022fedproto} introduced prototype sharing to alleviate label distribution heterogeneity without transmitting model gradients, yet it largely overlooks topological structures. To simultaneously capture structure and heterogeneity, FedPub~\cite{baek2023personalized} proposed functional embeddings for personalized subgraph training, and AdaFGL~\cite{li2024adafgl} along with FedGTA~\cite{li2024fedgta} integrated topology-confidence strategies to improve structure-aware aggregation. More recently, FedTAD~\cite{zhu2024fedtad} employed topology-aware, data-free distillation to achieve structure-sensitive aggregation, thereby reducing reliance on raw feature sharing.

Concurrently, standard FGL paradigms enforce a strict closed-world assumption, rendering them incapable of discovering novel categories that dynamically emerge across isolated clients. Integrating open-world discovery into FGL is therefore essential for robust deployment in real-world environments. Our proposed GCD-FGL framework is specifically designed to bridge this exact gap.

\section{Methodology}
\label{sec:methodology}

\begin{figure*}[H]
    \centering
    \includegraphics[width=1\linewidth]{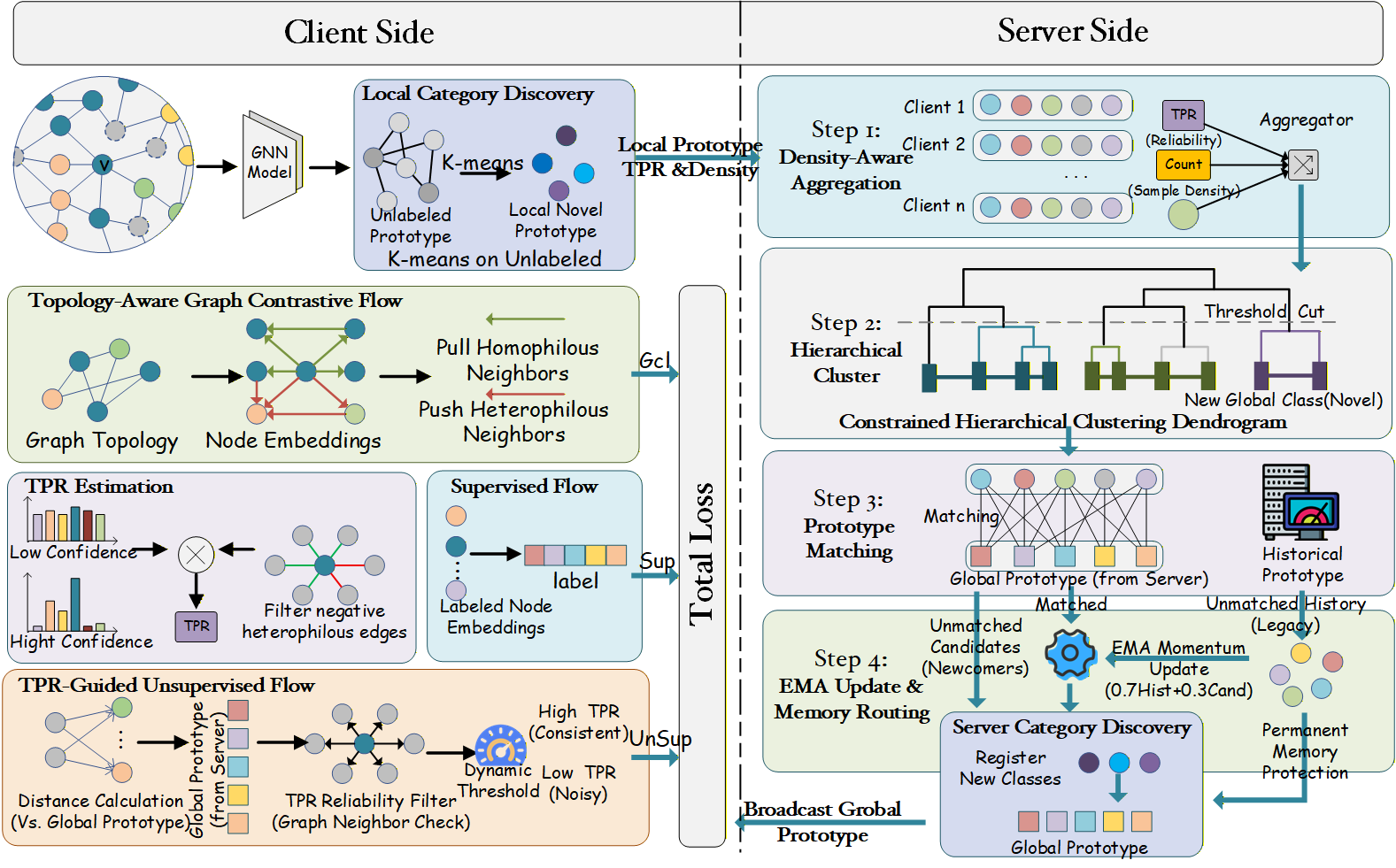}
    \caption{Overview of the proposed GCD-FGL framework.}
    \label{fig:framework}
\end{figure*}

\subsection{Overview of the GCD-FGL Framework}

The core principle of GCD-FGL is to treat global prototypes~\cite{snell2017prototypical} as the primary semantic bridge across clients, progressively refining these representations through reliability-aware local extraction and hierarchical global alignment. As illustrated in Figure~\ref{fig:framework}, unlike standard FGL frameworks~\cite{mcmahan2017communication} that are sensitive to subgraph distribution discrepancy and heterogeneous label spaces, GCD-FGL is driven by two systematically designed structural components:

\textbf{Topology-Reliable Semantic Alignment and Discovery (Client Side):} To mitigate the neighborhood absorption effect, we introduce a comprehensive alignment and discovery process guided by the TRG mechanism. This process filters noisy edges, directly mitigating the neighborhood absorption effect and preventing representation collapse for novel categories.

\textbf{Hierarchical Prototype Alignment (Server Side):} To resolve global semantic inconsistency across non-IID clients, the server performs Hierarchical Prototype Alignment to harmonize the global semantic space. By incorporating a Density-TPR weighting mechanism into the aggregation process, Hierarchical Prototype Alignment unifies decentralized knowledge and dynamically identifies novel categories without predefining the cluster count ($K$), effectively preventing amplified catastrophic forgetting.

\textbf{Collaborative Optimization Protocol.} The complete GCD-FGL training paradigm iteratively executes local topology-aware extraction and global hierarchical alignment. More specifically, the corresponding client-side local update procedure is detailed in Algorithm~\ref{alg:client_update}, while the overarching server-side aggregation and category discovery workflow is subsequently summarized in Algorithm~\ref{alg:gcd_fgl_server}.

\begin{algorithm}
\caption{GCD-FGL-Client}
\label{alg:client_update}
\begin{algorithmic}[1]
\Require Client ID $i$, global prototypes $\mathcal{P}_{global}$, global model $\omega^{(t-1)}$, local epochs $E$
\State Initialize local model $\omega_i \gets \omega^{(t-1)}$

\For{$e = 1,...,E$}
    \State Compute node embeddings $\tilde{Z}$ via local GNN
    
    \Statex \textit{TPR Estimation:}
    \State Compute TPR score $w_v$ for $v \in \mathcal{V}_i$ via Eq.~\ref{eq:tpr}
    
    \Statex \textit{Reliability-Aware Semantic Alignment:}
    \State Update $\omega_i$ by optimizing $\mathcal{L}_{total} = \mathcal{L}_{align} + \beta \mathcal{L}_{gcl}$ via Eq.~\ref{eq:ltotal}
\EndFor

\Statex \textit{Local Category Discovery \& Extraction:}
\State Extract local prototypes $\mathcal{P}_{i}^{known}$ and $\mathcal{P}_{i}^{novel}$ via clustering
\State Compute average TPR $w_i$ and the set of local sample densities $\mathbf{c}_i = \{c_{i,k}\}$
\State \Return $\{ \omega_i, \mathcal{P}_i, w_i, \mathbf{c}_i \}$, where $\mathcal{P}_i = \mathcal{P}_{i}^{known} \cup \mathcal{P}_{i}^{novel}$
\end{algorithmic}
\end{algorithm}

\begin{algorithm}
\caption{GCD-FGL-Server}
\label{alg:gcd_fgl_server}
\begin{algorithmic}[1]
\Require rounds $R$, momentum $\rho$, hierarchical penalty $\lambda_{hc}$

\For{each round $t = 1,...,R$}
    \State Sample a subset of clients $S_t$
    \For{each client $i \in S_t$ \textbf{in parallel}}
        \State $\{ \omega_i, \mathcal{P}_i, w_i, \mathbf{c}_i \} \gets \textsc{Client}(i, \mathcal{P}_{global}, \omega^{(t-1)})$
    \EndFor
    
    \Statex \textit{Step 1: Density-Aware Prototype Aggregation}
    \State Update global model via Eq.~\ref{eq:fedavg}
    \State Compute updated global prototypes $P_k^{(t)}$ via Eq.~\ref{eq:global_proto}
    
    \Statex \textit{Step 2: Hierarchical Prototype Alignment}
    \State Decompose each $\mathcal{P}_i$ into $\mathcal{P}_{i}^{known}$ and $\mathcal{P}_{i}^{novel}$ based on global known prototypes
\State Form global discovery pool $\mathcal{X}_{nov} = \bigcup_{i \in S_t} \mathcal{P}_{i}^{novel}$
\State Extract candidate novel centers $\mathcal{P}_{cand}$ via Eq.~\ref{eq:clustering}
    
    \Statex \textit{Step 3 \& 4: EMA Update \& Global Memory Routing}
    \State Update memory $\mathcal{P}_{global}$ with $\mathcal{P}_{cand}$ via Eq.~\ref{eq:ema}
    
    \State Broadcast $\omega^{(t)}$ and $\mathcal{P}_{global}$ to clients $S_t$
\EndFor
\end{algorithmic}
\end{algorithm}
\vspace{-0.5cm}

\subsection{Client Side: Topology-Reliable Semantic Alignment and Discovery}
\label{method:client}

\textbf{Motivation.} Structural truncation in isolated subgraphs exacerbates the neighborhood absorption effect, thereby inducing representation collapse for novel categories.
\textbf{Solution.} To address this issue, we propose a Topology-Reliable Semantic Alignment and Discovery process, which utilizes the \textbf{T}o\textbf{P}ology \textbf{R}eliability (TPR) metric to evaluate node trustworthiness to filter heterophilous noise.

\textbf{TPR Estimation.}
For a given node $v \in \mathcal{V}_i$ on client $i$, its predictive confidence $S_{conf}^{(v)}$ (distinguishing between states of High Confidence and Low Confidence) is derived from the normalized Shannon entropy over the predicted probability distribution $p_{v,c}$, a standard proxy for model uncertainty~\cite{grandvalet2005semi}:
\begin{equation}
S_{conf}^{(v)} = 1 - \frac{-\sum_{c=1}^{C} p_{v,c} \log(p_{v,c} + \epsilon)}{\max(\log C, \epsilon)}.
\label{eq:confidence}
\end{equation}

Concurrently, the structural smoothness $S_{homo}^{(v)}$ measures the semantic consistency between node $v$ and its immediate local neighborhood $\mathcal{N}_i(v)$:
\begin{equation}
S_{homo}^{(v)} = \frac{1}{\max(1, d_v)} \sum_{u \in \mathcal{N}_i(v)} \max(0, \tilde{z}_v^\top \tilde{z}_u).
\label{eq:smoothness}
\end{equation}

where $d_v$ is the local degree. The $\max(0, \cdot)$ operation explicitly functions to filter negative heterophilous edges. 

\textit{Intuition.} The final TPR score is designed as a strict logical conjunction to filter heterophilous noise, ensuring that a node influences subsequent alignments only if it simultaneously exhibits high predictive certainty ($S_{conf}$) and local semantic homophily ($S_{homo}$).
\begin{equation}
w_v = S_{homo}^{(v)} \cdot S_{conf}^{(v)}.
\label{eq:tpr}
\end{equation}

\textbf{Topology-Aware Graph Contrastive Flow.}
To capture intrinsic graph topology and mitigate representation collapse in structurally truncated regions~\cite{chuang2020debiased}, we employ a semantic graph contrastive loss ($\mathcal{L}_{gcl}$)~\cite{zhu2020deep}. Specifically, let $M_v \in \{0, 1\}$ denote a boolean mask indicating confident assignment ($M_v \equiv 1$ for labeled nodes), and $\tilde{y}_v$ denote the semantic label, which is the ground-truth $y_v$ for labeled nodes or the assigned pseudo-label $\hat{y}_v$ for unlabeled nodes. The adjusted positive similarity $s_{pos}(v,u)$ between connected nodes $(v,u) \in \mathcal{E}_i$ is formulated as follows:
\begin{equation}
s_{pos}(v,u) = \begin{cases} 10^{-8}, & \text{if } M_v \land M_u \text{ and } \tilde{y}_v \neq \tilde{y}_u \\ \exp\left(\frac{\tilde{z}_v^\top \tilde{z}_u}{\tau}\right), & \text{otherwise} \end{cases}.
\label{eq:spos}
\end{equation}

where setting the similarity to $10^{-8}$ avoids numerical instability in subsequent logarithmic calculations. The overall contrastive loss, which is symmetric for negative samples, is defined as follows.

\textit{Intuition.} Unlike conventional contrastive objectives that uniformly pull all connected nodes together, this topology-aware loss employs the false-positive truncation mechanism ($s_{pos}$) to nullify the pull of confident heterophilous edges.
\begin{equation}
\begin{split}
\mathcal{L}_{gcl} = & \frac{1}{|\mathcal{V}_i|} \sum_{v \in \mathcal{V}_i} \frac{1}{d_v} \sum_{u \in \mathcal{N}_i(v)} \\
& -\log \left( \frac{s_{pos}(v,u)}{s_{pos}(v,u) + \sum_{u'} s_{neg}(v,u')} \right).
\end{split}
\label{eq:lgcl}
\end{equation}

where $s_{neg}(v,u') = \exp\left(\frac{\tilde{z}_v^\top \tilde{z}_{u'}}{\tau}\right)$ for negative samples $u' \notin \mathcal{N}_i(v)$, following the standard InfoNCE formulation.

\textbf{Supervised Flow.}
For nodes with ground-truth labels $\mathcal{V}_i^L$, the supervised flow employs a cross-entropy objective to align local embeddings against the global prototype buffer $\mathcal{P}_{global}$ broadcasted by the server:
\begin{equation}
\mathcal{L}_{sup} = \frac{1}{|\mathcal{V}_i^L|} \sum_{v \in \mathcal{V}_i^L} \text{CE} \left( \frac{\tilde{z}_v \mathcal{P}_{global}^\top}{\tau}, y_v \right).
\label{eq:lsup}
\end{equation}

where $\text{CE}(\cdot, \cdot)$ denotes the cross-entropy loss applied to the softmax-normalized logits.

\textbf{TPR-Guided Unsupervised Flow.}
For unlabeled nodes $\mathcal{V}_i^U$, we generate pseudo-labels $\hat{y}_v$ using a sharpened temperature $\tau_{sharp}$ over the currently available global prototypes in $\mathcal{P}_{global}$. To filter out noisy assignments, a dynamic threshold $\gamma$ is calculated based on batch confidence statistics:
\begin{equation}
\gamma = \max(0.5, \mu_q + \alpha \cdot \sigma_q).
\label{eq:gamma}
\end{equation}

where $\mu_q$ and $\sigma_q$ are the mean and standard deviation of prediction confidences within the current batch. Unlike fixed thresholds that may discard too many samples early in training~\cite{sohn2020fixmatch}, this batch-adaptive formulation dynamically scales with the model's learning pace. The boolean mask is formally defined as $M_v = \mathbb{I}(q_v > \gamma)$ to retain only reliable predictions. 

\textit{Intuition.} To robustly align with the global prototype, this loss employs two safety mechanisms: the adaptive mask ($M_v$) maintains a steady learning curriculum, while the TPR score ($w_v$) acts as a soft weight to suppress gradients from surviving heterophilous noise.
\begin{equation}
\mathcal{L}_{unsup} = \frac{1}{|\mathcal{V}_i^U|} \sum_{v \in \mathcal{V}_i^U} \left[ w_v \cdot M_v \cdot \text{CE} \left( \frac{\tilde{z}_v \mathcal{P}_{global}^\top}{\tau}, \hat{y}_v \right) \right].
\label{eq:lunsup}
\end{equation}

\textbf{Total Loss.}
The Total Loss optimized iteratively by client $i$ balances supervised discrimination, unsupervised alignment, and structural consistency:
\begin{equation}
\mathcal{L}_{total} = \underbrace{\mathcal{L}_{sup} + \mathcal{L}_{unsup}}_{\mathcal{L}_{align}} + \beta \mathcal{L}_{gcl}.
\label{eq:ltotal}
\end{equation}

Upon convergence, the client extracts known ($\mathcal{P}_{i}^{known}$) and novel ($\mathcal{P}_{i}^{novel}$) prototypes and their sample densities via local K-means on unlabeled data, sending them to the server.

\textbf{Local Category Discovery.} 
In the GCD setting, the true number of novel categories is unknown. Instead of a predefined $K$, we employ relaxed over-clustering and density-filtering. For the local K-means, the relaxed local cluster count $K_{local}$ is dynamically upper-bounded by the unlabeled graph volume to prevent over-fragmentation:
\begin{equation}
K_{local} = \max\left(2, \min\left(K_{max}, \lfloor \frac{|\mathcal{V}_i^U|}{3} \rfloor \right)\right).
\label{eq:k_local}
\end{equation}

where $K_{max}$ is an empirical upper bound for the expected number of novel categories. Equation~\ref{eq:k_local} serves as a relaxed upper bound to encourage exploration. Consequently, in edge cases, such as a very small $|\mathcal{V}_i^U|$ or when unlabeled nodes exclusively belong to known categories, this relaxation might propose spurious boundaries. However, the subsequent density filters out these false novel prototypes by accepting a local cluster $C_k$ only if it satisfies a minimum sample density constraint $|C_k| > \tau_{density}$ with $\tau_{density}=5$. Finally, server-side hierarchical clustering structurally corrects any surviving spurious prototypes to mitigate false category generation.

\subsection{Server Side: Hierarchical Category Discovery}
\label{method:server}

\textbf{Motivation.} The local representation collapse often propagates to the server, interacting with label skew to drive and amplify global semantic inconsistency across clients. 
\textbf{Solution.} We design a Hierarchical Prototype Alignment strategy to identify novel category candidates and maintain memory stability.

The server performs a progressive semantic association process, where reliable local discoveries are first aggregated and subsequently structurally grouped, before being finally stabilized through temporal memory.

\textbf{Step 1: Density-Aware Aggregation.}
To mitigate label skew, the server computes the updated global prototype $P_k^{(t)}$ using a joint weight $v_{i,k} = w_{i,k} \cdot c_{i,k}$ (where $w_{i,k}$ represents the cluster-specific average TPR score for local cluster $k$ and $c_{i,k} \in \mathbf{c}_i$ represents the local Sample Density for category $k$) within the central Aggregator:
\begin{equation}
P_k^{(t)} = \text{Normalize} \left( \frac{\sum_{i} v_{i,k} P_{i,k}}{\sum_{i} v_{i,k} + \epsilon} \right).
\label{eq:global_proto}
\end{equation}

This joint weighting mitigates weight-shifting from non-IID distributions, preventing noisy clients from dominating the global space. Here, $P_{i,k} \in \mathcal{P}_i$ denotes the local prototype for category $k$ received from client $i$.

\textbf{Step 2: Hierarchical Cluster for Server Category Discovery.}
Local known category prototypes $\mathcal{P}_{i}^{known}$ are absorbed during Step 1 aggregation. Therefore, only the unabsorbed local novel prototypes $\mathcal{P}_{i}^{novel}$ are injected into the global discovery pool $\mathcal{X}_{nov} = \bigcup_{i \in S_t} \mathcal{P}_{i}^{novel}$. To extract global candidate centers $\mathcal{P}_{cand}$ for novel categories, we introduce a penalized Silhouette score~\cite{rousseeuw1987silhouettes} to find the optimal Threshold Cut $\theta^*$ on the Constrained Hierarchical Clustering Dendrogram.

\textit{Intuition.} Explicitly penalizing the cluster count enforces compact, meaningful semantic groupings and rejects fragmented boundaries from isolated clients.
\begin{equation}
\theta^* = \arg\max_{\theta} \left( S_{sil}(\theta) - \lambda_{hc} \cdot \Omega(\theta) \right).
\label{eq:clustering}
\end{equation}

where $\Omega(\theta) = \max(0, N_{clust}(\theta) - (|\mathcal{Y}_{known}| + 2))$ represents the hierarchical penalty, and $\lambda_{hc}$ is the regularization factor.

\textbf{Step 3: Prototype Matching.} 
To match candidate centers $\mathcal{P}_{cand}$ with historical prototypes $\mathcal{P}_{hist}^{(t-1)}$, we compute the cosine similarity matrix $S \in \mathbb{R}^{|\mathcal{P}_{cand}| \times |\mathcal{P}_{hist}|}$, where $S_{r,c} = \cos(P_{cand, r}, P_{hist, c}^{(t-1)})$. The optimal assignment is solved via the Hungarian algorithm to maximize the total similarity:
\begin{equation}
\arg\max_{A} \sum_{r,c} A_{r,c} \cdot S_{r,c}, \quad \text{s.t.} \sum_r A_{r,c} \le 1, \sum_c A_{r,c} \le 1.
\label{eq:matching}
\end{equation}

where $A_{r,c} \in \{0, 1\}$ is the boolean assignment matrix. To reject mismatches, a match is considered valid only if the similarity exceeds a dynamic threshold: $S_{r,c} > \max(\tau_{base}, \bar{S})$, where $\bar{S}$ is the mean similarity of the current matrix, and $\tau_{base}$ is the base semantic similarity threshold empirically set to 0.3 across all datasets.

\textbf{Step 4: EMA Update \& Memory Routing.}
Valid matched pairs undergo an EMA update for temporal smoothing with the corresponding historical prototype $P_{hist, c}^{(t-1)}$, controlled by a momentum factor $\rho$:
\begin{equation}
P_{hist, c}^{(t)} = \text{Normalize} \left( \rho \cdot P_{hist, c}^{(t-1)} + (1 - \rho) \cdot P_{cand, r} \right).
\label{eq:ema}
\end{equation}

where $r$ is the specific candidate index matched to the historical prototype $c$ via the aforementioned assignment matrix $A$. Unmatched candidates are registered as new additions to the global prototype buffer $\mathcal{P}_{global}$, while unmatched historical prototypes are preserved to prevent forgetting.

\section{Experiments}
\label{Experiments}

\begin{table*}[htbp]
    \centering
    \caption{Statistical information and GCD split protocol of the experimental datasets.}
    \label{table:datasetSplit}
    \resizebox{\textwidth}{!}{
    \begin{tabular}{l|l rrr |cc c}
    \toprule
        \multirow{2}{*}{\textbf{Dataset}} & \multirow{2}{*}{\textbf{Description}} & \multirow{2}{*}{\textbf{Nodes}} & \multirow{2}{*}{\textbf{Edges}} & \multirow{2}{*}{\textbf{Features}} & \multicolumn{2}{c}{\textbf{categories}} & \multirow{2}{*}{\textbf{Train/Val/Test*}} \\ \cmidrule(lr){6-7}
        & & & & & \textbf{Total} & \textbf{Known / Novel} & \\ \midrule
        Cora & Citation network & 2,708 & 5,429 & 1,433 & 7 & 4 / 3 & 20\% / 40\% / 40\% \\
        CiteSeer & Citation network & 3,327 & 4,732 & 3,703 & 6 & 3 / 3 & 20\% / 40\% / 40\% \\
        Amazon Photo & Co-purchase graph & 7,487 & 119,043 & 745 & 8 & 4 / 4 & 20\% / 40\% / 40\% \\
        Amazon Computers & Co-purchase graph & 13,381 & 245,778 & 767 & 10 & 5 / 5 & 20\% / 40\% / 40\% \\
        Coauthor CS & Co-authorship graph & 18,333 & 81,894 & 6,805 & 15 & 8 / 7 & 20\% / 40\% / 40\% \\ \bottomrule
    \end{tabular}}
    \vspace{1ex}
    \raggedright \footnotesize \textit{*Note: The 20\%/40\%/40\% (Train/Val/Test) split applies only to Known categories; Novel categories remain entirely unlabeled.}
\end{table*}

In this section, we conduct extensive experiments to evaluate the effectiveness of the proposed GCD-FGL framework. We first introduce the graph datasets, describe the GCD scenarios, and outline the baselines and evaluation metrics. Subsequently, we aim to answer the following research questions:
\textbf{Q1 (Performance Comparison):} Compared with other state-of-the-art baselines, can GCD-FGL achieve better performance in GCD scenarios?
\textbf{Q2 (Ablation Study):} Where does the performance gain of GCD-FGL come from?
\textbf{Q3 (Hyperparameter Sensitivity):} How sensitive is GCD-FGL to the variations of key hyperparameters?
\textbf{Q4 (Robustness):} How robust is the proposed GCD-FGL method?
\textbf{Q5 (Efficiency):} How does GCD-FGL perform in terms of training efficiency and computational cost?

\subsection{Experiments Setup}

\textbf{Datasets and GCD setup.} To evaluate the effectiveness of GCD-FGL, we conduct experiments on five widely used graph benchmark datasets: Cora, CiteSeer, Amazon Photo, Amazon Computers, and Coauthor CS. To simulate a realistic federated GCD environment, we employ the Louvain algorithm~\cite{blondel2008fast} to partition the global graph topology into 10 clients. 

As summarized in Table~\ref{table:datasetSplit}, our proposed two-stage GCD split protocol operates as follows. First, in the \textit{Global Category Split}, we partition the global label space into disjoint Known and Novel categories. For datasets with an odd number of total categories $|\mathcal{Y}|$, the number of Known categories is set to $\lceil |\mathcal{Y}|/2 \rceil$. Second, in the \textit{Local Data Masking} stage, we construct the dataset masks within each client's isolated subgraph under a transductive setting. Specifically, for Known categories, 20\% of the instances are sampled to form the labeled training set, 40\% are allocated to the validation set, and the remaining 40\% are retained as unlabeled data. These unlabeled Known instances are then combined with 100\% of the Novel category instances to concurrently form the unlabeled training set and the testing set, requiring the model to discover new categories across all local unlabeled data.

\textbf{Baselines.} We evaluate GCD-FGL against representative state-of-the-art baselines, which are broadly categorized into three groups: adapted computer vision GCD methods, adapted graph GCD methods, and dedicated federated GCD methods. For the adapted CV GCD methods, we select AutoNovel~\cite{han2020autonovel}, ORCA~\cite{cao2022open}, GCD~\cite{vaze2022generalized}, and SimGCD~\cite{wen2023parametric}. Following standard adaptation protocols, we extend these centralized models to the federated graph setting by equipping them with a GCN backbone~\cite{kipf2016semi} and employing standard FedAvg~\cite{mcmahan2017communication} for server-side aggregation. For the adapted graph GCD methods, we evaluate SWIRL~\cite{deng2025towards}, a state-of-the-art approach tailored for holistic GGCD, which is similarly extended to the federated setting via FedAvg. Finally, to ensure a comprehensive comparison against architectures explicitly designed for distributed scenarios, we benchmark against FedGCD~\cite{pu2024federated}, a recent framework formulated specifically for federated generalized category discovery.

\textbf{Evaluation Metrics.} Following standard GCD evaluation protocols, we adopt a comprehensive metric system. Since the cluster assignments of novel categories are permutation-invariant, the Hungarian algorithm~\cite{kuhn1955hungarian} is applied to find the optimal matching between the predicted cluster assignments and the ground-truth labels before computing accuracy. In line with the realistic GCD setting, the total number of clusters ($K$) is not provided as an oracle prior. The evaluated metrics include New Acc (accuracy on novel nodes), Old Acc (accuracy on known nodes), All Acc (overall accuracy on the entire test set), and HRScore. It is important to clarify that All Acc is evaluated independently over the entire test set and is not a direct mathematical derivation of the Old and New accuracies. Specifically, in alignment with recent Graph GCD studies like SWIRL~\cite{deng2025towards}, HRScore is defined as the harmonic mean of Old Acc and New Acc ($2 \times (\text{Old Acc} \times \text{New Acc}) / (\text{Old Acc} + \text{New Acc})$). Because relying solely on overall accuracy, known-category accuracy, or novel-category accuracy in isolation cannot reflect a model's comprehensive capability, we adopt HRScore as the primary evaluation metric to quantify the balance between old-knowledge retention and new-category exploration.

\textbf{Implementation Details.} All experiments and federated framework simulations are implemented based on the OpenFGL library~\cite{li2024openfgl}. We consistently utilize a 2-layer GCN~\cite{kipf2016semi} as the backbone architecture. Models are optimized using the Adam optimizer~\cite{kingma2014adam} with an initial global learning rate of 0.001 and a weight decay of 5e-4, while the temperature parameter for contrastive learning is set to 0.1. For baseline comparisons, all methods are executed using the default hyperparameters specified in their original papers. Since this work is the first to explore GCD tasks under a federated graph learning challenge, there are no existing Fed-GCD-specific methods for direct comparison. To adapt existing centralized and non-graph GCD algorithms to the federated environment, we implement a standard federated baseline protocol. For visual baselines, we replace image augmentations with standard graph contrastive learning via edge dropping and feature masking~\cite{zhu2020deep}. Specifically, during training, all baseline models employ FedAvg for model weight aggregation. During evaluation, to enable cross-client category discovery and ensure a fair comparison against our framework, we implement a naive global prototype aggregation mechanism on the server. This mechanism aligns local prototypes from isolated clients to global prototypes via Hungarian matching based on cosine distance, followed by simple averaging. This uniform baseline adaptation ensures that all models are evaluated under the exact same federated prototype-matching conditions. All experiments are conducted on an Ubuntu 22.04 workstation equipped with an Intel Core i9-13900K processor, an NVIDIA GeForce RTX 3090 GPU (24GB), CUDA 12.1, and 64GB of RAM. To ensure the reliability of our findings, we report the mean performance and standard deviation across multiple independent runs.

\subsection{Performance Comparison}

\begin{table*}
\centering
\caption{Performance comparison across five graph datasets. The best results are highlighted in bold, and the second-best results are underlined. The `Gain` column shows the improvement between GCD-FGL and the best baseline.}
\label{tab:main_results}
\resizebox{\textwidth}{!}{
\begin{tabular}{l|l|ccccccc|c}
\toprule
Datasets & Metrics & AutoNovel & ORCA & GCD & SimGCD & FedGCD & SWIRL & GCD-FGL & Gain \\
\midrule
\multirow{4}{*}{Cora} 
& \cellcolor{gray!10}HRScore & \cellcolor{gray!10}$36.98_{\scriptstyle\,\pm\,0.21}$ & \cellcolor{gray!10}$34.40_{\scriptstyle\,\pm\,0.53}$ & \cellcolor{gray!10}$37.75_{\scriptstyle\,\pm\,3.59}$ & \cellcolor{gray!10}$32.99_{\scriptstyle\,\pm\,0.17}$ & \cellcolor{gray!10}$37.44_{\scriptstyle\,\pm\,6.79}$ & \cellcolor{gray!10}$\underline{48.96}_{\scriptstyle\,\pm\,0.86}$ & \cellcolor{gray!10}$\mathbf{52.85_{\scriptstyle\,\pm\,1.24}}$ & \cellcolor{gray!10}+3.89 \\
\cmidrule{2-10}
& Old  & $\underline{58.01}_{\scriptstyle\,\pm\,0.57}$ & $41.67_{\scriptstyle\,\pm\,0.26}$ & $42.63_{\scriptstyle\,\pm\,8.88}$ & $54.17_{\scriptstyle\,\pm\,0.26}$ & $52.24_{\scriptstyle\,\pm\,16.17}$ & $51.60_{\scriptstyle\,\pm\,1.72}$ & $\mathbf{66.67_{\scriptstyle\,\pm\,1.44}}$ & +8.66 \\
& New  & $27.14_{\scriptstyle\,\pm\,0.19}$ & $29.29_{\scriptstyle\,\pm\,0.76}$ & $33.88_{\scriptstyle\,\pm\,1.41}$ & $23.72_{\scriptstyle\,\pm\,0.17}$ & $29.18_{\scriptstyle\,\pm\,6.53}$ & $\mathbf{46.58_{\scriptstyle\,\pm\,0.66}}$ & $\underline{43.78}_{\scriptstyle\,\pm\,1.58}$ & -2.80 \\
& All  & $31.38_{\scriptstyle\,\pm\,0.57}$ & $30.99_{\scriptstyle\,\pm\,0.65}$ & $35.08_{\scriptstyle\,\pm\,1.72}$ & $27.90_{\scriptstyle\,\pm\,0.14}$ & $32.35_{\scriptstyle\,\pm\,3.45}$ & $\underline{46.27}_{\scriptstyle\,\pm\,0.81}$ & $\mathbf{47.92_{\scriptstyle\,\pm\,1.28}}$ & +1.65 \\
\midrule
\multirow{4}{*}{CiteSeer} 
& \cellcolor{gray!10}HRScore & \cellcolor{gray!10}$44.62_{\scriptstyle\,\pm\,0.57}$ & \cellcolor{gray!10}$49.01_{\scriptstyle\,\pm\,0.31}$ & \cellcolor{gray!10}$38.81_{\scriptstyle\,\pm\,1.81}$ & \cellcolor{gray!10}$40.78_{\scriptstyle\,\pm\,0.44}$ & \cellcolor{gray!10}$32.01_{\scriptstyle\,\pm\,1.64}$ & \cellcolor{gray!10}$\underline{57.16}_{\scriptstyle\,\pm\,1.57}$ & \cellcolor{gray!10}$\mathbf{59.78_{\scriptstyle\,\pm\,1.08}}$ & \cellcolor{gray!10}+2.62 \\
\cmidrule{2-10}
& Old  & $52.78_{\scriptstyle\,\pm\,0.83}$ & $\mathbf{66.07_{\scriptstyle\,\pm\,0.40}}$ & $38.42_{\scriptstyle\,\pm\,0.63}$ & $37.70_{\scriptstyle\,\pm\,0.75}$ & $59.07_{\scriptstyle\,\pm\,1.87}$ & $55.66_{\scriptstyle\,\pm\,2.21}$ & $\underline{60.73}_{\scriptstyle\,\pm\,2.09}$ & -5.34 \\
& New  & $38.64_{\scriptstyle\,\pm\,0.73}$ & $38.95_{\scriptstyle\,\pm\,0.37}$ & $39.20_{\scriptstyle\,\pm\,3.63}$ & $44.41_{\scriptstyle\,\pm\,0.15}$ & $21.95_{\scriptstyle\,\pm\,1.52}$ & $\underline{58.75}_{\scriptstyle\,\pm\,2.21}$ & $\mathbf{58.85_{\scriptstyle\,\pm\,0.72}}$ & +0.10 \\
& All  & $41.77_{\scriptstyle\,\pm\,0.30}$ & $44.95_{\scriptstyle\,\pm\,0.20}$ & $39.03_{\scriptstyle\,\pm\,2.72}$ & $42.93_{\scriptstyle\,\pm\,0.20}$ & $30.17_{\scriptstyle\,\pm\,1.03}$ & $\underline{58.07}_{\scriptstyle\,\pm\,1.24}$ & $\mathbf{59.94_{\scriptstyle\,\pm\,0.81}}$ & +1.87 \\
\midrule
\multirow{4}{*}{CS} 
& \cellcolor{gray!10}HRScore & \cellcolor{gray!10}$57.77_{\scriptstyle\,\pm\,2.37}$ & \cellcolor{gray!10}$59.34_{\scriptstyle\,\pm\,8.28}$ & \cellcolor{gray!10}$56.01_{\scriptstyle\,\pm\,2.34}$ & \cellcolor{gray!10}$54.53_{\scriptstyle\,\pm\,0.14}$ & \cellcolor{gray!10}$53.63_{\scriptstyle\,\pm\,17.45}$ & \cellcolor{gray!10}$\underline{63.94}_{\scriptstyle\,\pm\,3.22}$ & \cellcolor{gray!10}$\mathbf{72.30_{\scriptstyle\,\pm\,6.63}}$ & \cellcolor{gray!10}+8.36 \\
\cmidrule{2-10}
& Old  & $62.06_{\scriptstyle\,\pm\,5.38}$ & $79.87_{\scriptstyle\,\pm\,28.21}$ & $67.74_{\scriptstyle\,\pm\,6.01}$ & $56.37_{\scriptstyle\,\pm\,0.12}$ & $41.31_{\scriptstyle\,\pm\,20.65}$ & $\underline{82.19}_{\scriptstyle\,\pm\,10.39}$ & $\mathbf{82.38_{\scriptstyle\,\pm\,17.01}}$ & +0.19 \\
& New  & $54.04_{\scriptstyle\,\pm\,0.77}$ & $47.21_{\scriptstyle\,\pm\,3.56}$ & $47.75_{\scriptstyle\,\pm\,1.63}$ & $52.80_{\scriptstyle\,\pm\,0.24}$ & $\mathbf{76.41_{\scriptstyle\,\pm\,5.41}}$ & $52.32_{\scriptstyle\,\pm\,0.96}$ & $\underline{64.41}_{\scriptstyle\,\pm\,2.48}$ & -12.00 \\
& All  & $55.15_{\scriptstyle\,\pm\,0.33}$ & $51.74_{\scriptstyle\,\pm\,0.94}$ & $50.52_{\scriptstyle\,\pm\,1.37}$ & $53.30_{\scriptstyle\,\pm\,0.23}$ & $\underline{66.55}_{\scriptstyle\,\pm\,3.80}$ & $56.45_{\scriptstyle\,\pm\,0.73}$ & $\mathbf{71.03_{\scriptstyle\,\pm\,0.33}}$ & +4.48 \\
\midrule
\multirow{4}{*}{Photo} 
& \cellcolor{gray!10}HRScore & \cellcolor{gray!10}$48.49_{\scriptstyle\,\pm\,0.55}$ & \cellcolor{gray!10}$42.54_{\scriptstyle\,\pm\,4.32}$ & \cellcolor{gray!10}$47.66_{\scriptstyle\,\pm\,3.77}$ & \cellcolor{gray!10}$37.68_{\scriptstyle\,\pm\,0.69}$ & \cellcolor{gray!10}$45.79_{\scriptstyle\,\pm\,5.21}$ & \cellcolor{gray!10}$\underline{48.61}_{\scriptstyle\,\pm\,2.86}$ & \cellcolor{gray!10}$\mathbf{51.66_{\scriptstyle\,\pm\,2.14}}$ & \cellcolor{gray!10}+3.05 \\
\cmidrule{2-10}
& Old  & $45.78_{\scriptstyle\,\pm\,0.68}$ & $38.35_{\scriptstyle\,\pm\,6.88}$ & $43.30_{\scriptstyle\,\pm\,5.34}$ & $33.78_{\scriptstyle\,\pm\,0.80}$ & $42.22_{\scriptstyle\,\pm\,8.03}$ & $\underline{62.27}_{\scriptstyle\,\pm\,5.04}$ & $\mathbf{62.29_{\scriptstyle\,\pm\,4.30}}$ & +0.02 \\
& New  & $\underline{51.53}_{\scriptstyle\,\pm\,0.91}$ & $47.75_{\scriptstyle\,\pm\,2.17}$ & $\mathbf{53.00_{\scriptstyle\,\pm\,4.77}}$ & $42.61_{\scriptstyle\,\pm\,1.22}$ & $50.01_{\scriptstyle\,\pm\,5.25}$ & $39.87_{\scriptstyle\,\pm\,3.24}$ & $44.13_{\scriptstyle\,\pm\,2.25}$ & -8.87 \\
& All  & $49.83_{\scriptstyle\,\pm\,0.84}$ & $44.97_{\scriptstyle\,\pm\,0.97}$ & $\underline{50.13}_{\scriptstyle\,\pm\,3.21}$ & $40.00_{\scriptstyle\,\pm\,0.89}$ & $47.71_{\scriptstyle\,\pm\,2.92}$ & $46.49_{\scriptstyle\,\pm\,2.40}$ & $\mathbf{50.19_{\scriptstyle\,\pm\,1.26}}$ & +0.06 \\
\midrule
\multirow{4}{*}{Computers} 
& \cellcolor{gray!10}HRScore & \cellcolor{gray!10}$42.17_{\scriptstyle\,\pm\,4.70}$ & \cellcolor{gray!10}$33.77_{\scriptstyle\,\pm\,5.53}$ & \cellcolor{gray!10}$46.21_{\scriptstyle\,\pm\,2.32}$ & \cellcolor{gray!10}$\underline{48.26}_{\scriptstyle\,\pm\,1.24}$ & \cellcolor{gray!10}$46.07_{\scriptstyle\,\pm\,3.05}$ & \cellcolor{gray!10}$40.87_{\scriptstyle\,\pm\,0.23}$ & \cellcolor{gray!10}$\mathbf{54.62_{\scriptstyle\,\pm\,4.49}}$ & \cellcolor{gray!10}+6.36 \\
\cmidrule{2-10}
& Old  & $\underline{56.10}_{\scriptstyle\,\pm\,2.74}$ & $52.75_{\scriptstyle\,\pm\,4.51}$ & $45.61_{\scriptstyle\,\pm\,3.84}$ & $51.70_{\scriptstyle\,\pm\,1.68}$ & $45.02_{\scriptstyle\,\pm\,3.06}$ & $35.02_{\scriptstyle\,\pm\,0.28}$ & $\mathbf{58.39_{\scriptstyle\,\pm\,6.83}}$ & +2.29 \\
& New  & $33.78_{\scriptstyle\,\pm\,5.95}$ & $24.83_{\scriptstyle\,\pm\,5.90}$ & $46.83_{\scriptstyle\,\pm\,2.52}$ & $45.25_{\scriptstyle\,\pm\,1.67}$ & $47.17_{\scriptstyle\,\pm\,5.43}$ & $\underline{49.08}_{\scriptstyle\,\pm\,0.36}$ & $\mathbf{51.30_{\scriptstyle\,\pm\,5.86}}$ & +2.22 \\
& All  & $45.36_{\scriptstyle\,\pm\,1.82}$ & $39.31_{\scriptstyle\,\pm\,2.33}$ & $46.20_{\scriptstyle\,\pm\,1.55}$ & $\underline{51.50}_{\scriptstyle\,\pm\,0.39}$ & $46.05_{\scriptstyle\,\pm\,3.30}$ & $41.79_{\scriptstyle\,\pm\,0.52}$ & $\mathbf{53.03_{\scriptstyle\,\pm\,3.02}}$ & +1.53 \\
\bottomrule
\end{tabular}
}
\end{table*}

To answer \textbf{Q1}, we evaluate the performance of GCD-FGL against baseline models across five benchmark graph datasets under the distributed federated setting (partitioned via the Louvain algorithm). Detailed results are presented in Table~\ref{tab:main_results}, with the best results highlighted in bold. Measured by the HRScore, which quantifies the balance between known category retention and novel category discovery, our proposed method consistently outperforms the baselines across all five datasets. Most notably, GCD-FGL achieves an average gain of +4.86 points in HRScore over the strongest competitors and secures the highest overall accuracy (All Acc) across all benchmarks. For instance, on the Coauthor CS dataset, GCD-FGL yields an HRScore of 72.30\%, an improvement of 8.36\% over the SWIRL baseline (63.94\%).

While GCD-FGL achieves state-of-the-art comprehensive performance, it occasionally yields to specific baselines on isolated metrics. For instance, on the Cora dataset, SWIRL achieves a marginally higher New Acc (46.58\%) compared to GCD-FGL (43.78\%). However, in FGGCD settings, aggressively optimizing for novel category discovery often induces feature space squeezing, inherently compromising known category retention. Consequently, SWIRL's slight gain in New Acc comes at the expense of its Old Acc (51.60\%, compared to GCD-FGL's 66.67\%). This trade-off emphasizes the critical importance of utilizing HRScore as the primary evaluation metric, as it demonstrates that GCD-FGL achieves a superior harmonic balance without suffering from catastrophic forgetting of existing knowledge.

The results also reveal performance fluctuations in existing baselines, particularly concerning known category retention (Old Acc). For instance, FedGCD (on the CS dataset) and SWIRL (on the Computers dataset) experience notable accuracy drops on known categories, occasionally falling below their respective performance on novel categories. This vulnerability stems from their respective architectural limitations. Although SWIRL is tailored for GGCD, it lacks mechanisms to address the subgraph distribution discrepancy and heterogeneous label spaces inherent in federated settings, thereby hindering its generalization across distributed clients. Conversely, while FedGCD explicitly accounts for federated heterogeneity, its core logic, originally designed for visual tasks, largely overlooks complex topological dependencies. Consequently, it is susceptible to structural truncation, limiting its convergence and representation quality, particularly on sparse datasets such as Cora and CiteSeer.

\subsection{Method Interpretability}

\begin{figure}
    \centering
    \includegraphics[width=1\linewidth]{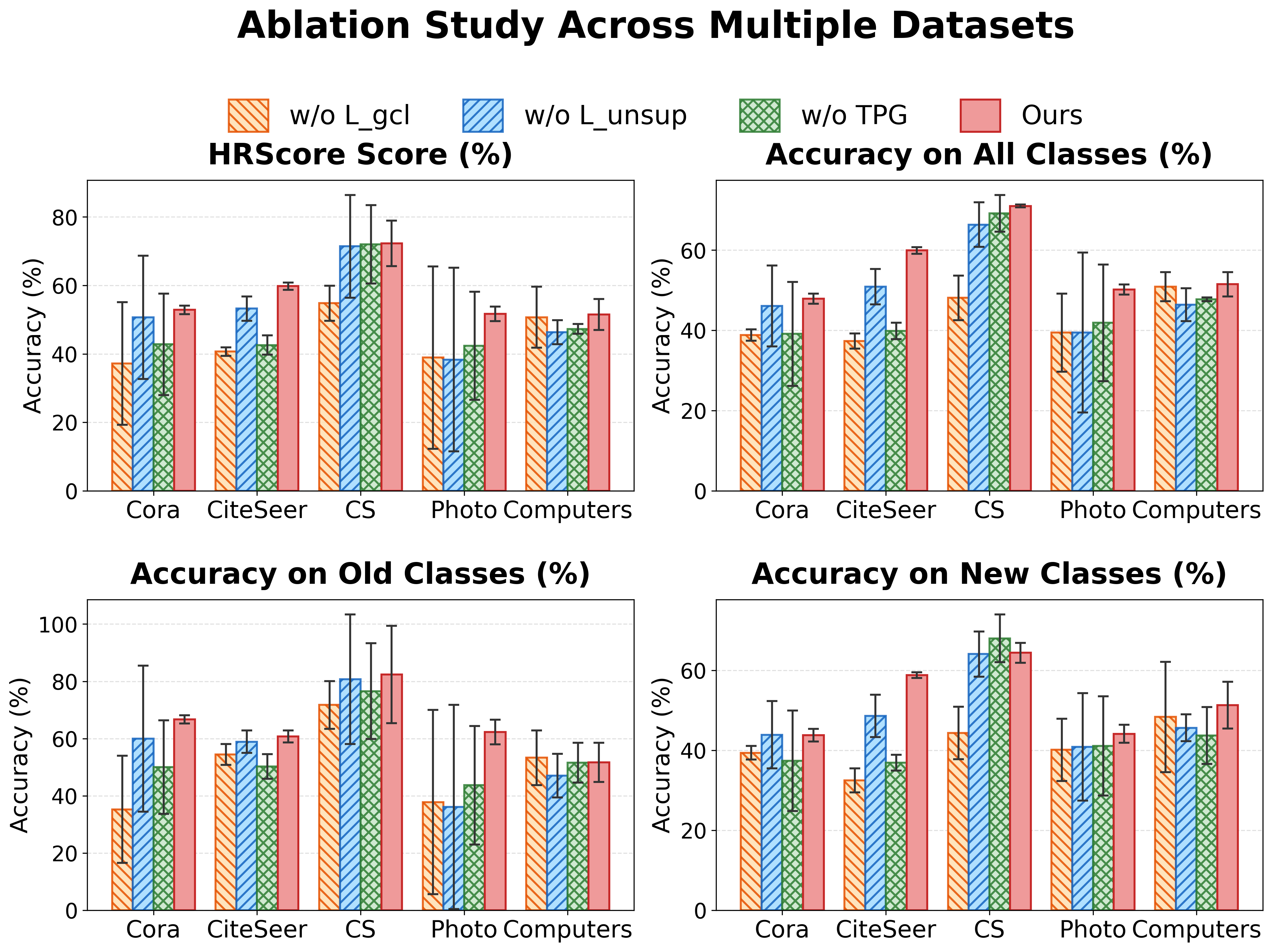}
    \caption{Quantitative ablation study across all five datasets. The error bars indicate performance variance.}
    \label{fig:ablation_bars}
\end{figure}

\begin{figure}
    \centering
    \includegraphics[width=1\linewidth]{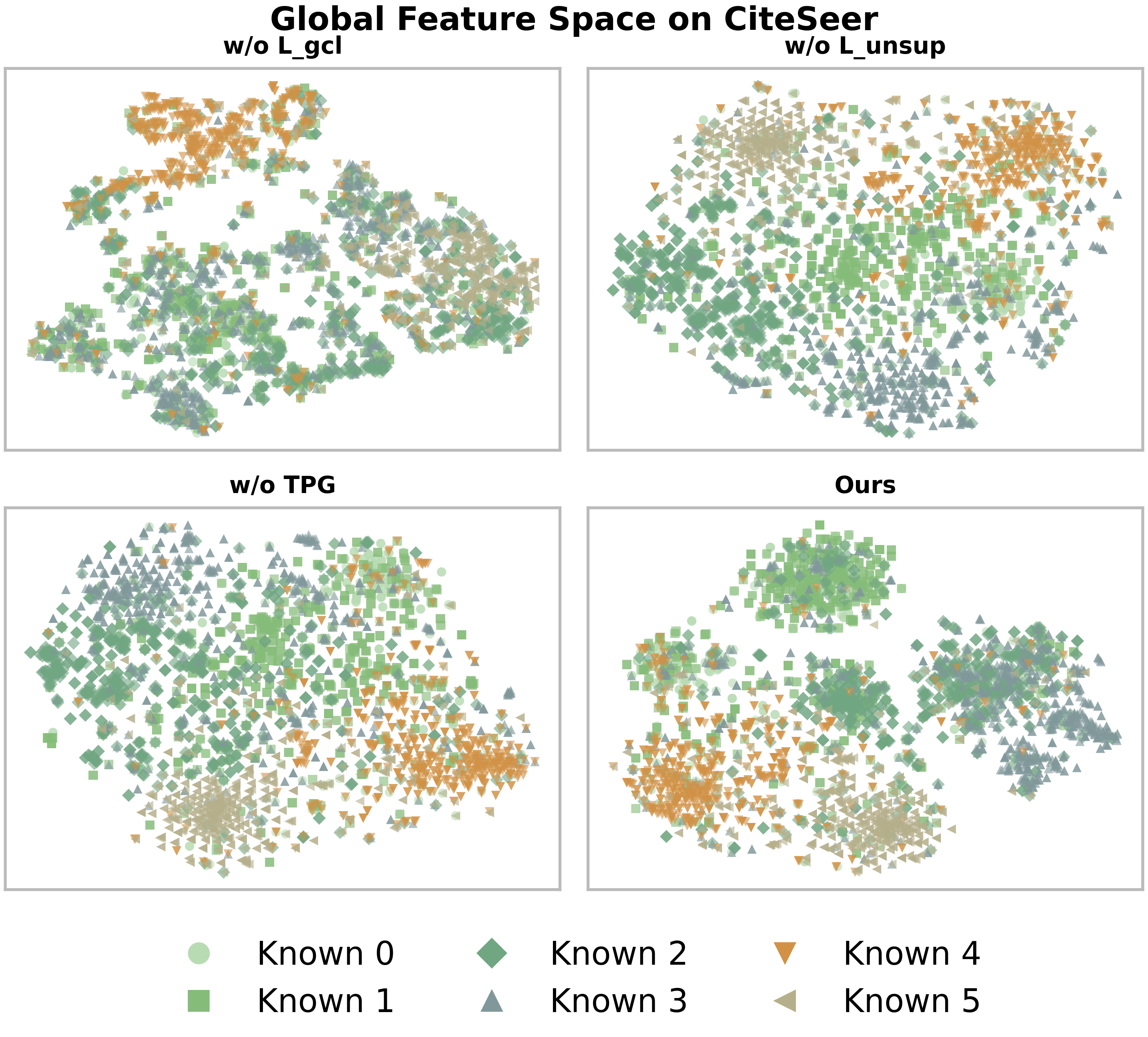}
    \caption{t-SNE visualization of node feature distributions for different ablated variants on the CiteSeer dataset.}
    \label{fig:tsne_citeseer}
\end{figure}

To answer \textbf{Q2}, we conduct an ablation study to systematically isolate and evaluate the individual contributions of the core components within GCD-FGL: the TPR-Guided Unsupervised Flow ($\mathcal{L}_{unsup}$), the Topology-Aware Graph Contrastive Flow ($\mathcal{L}_{gcl}$), and the TRG mechanism. This evaluation is performed under a standard distributed setting (partitioned via the Louvain algorithm). Quantitative results across all five datasets are comprehensively detailed in Figure~\ref{fig:ablation_bars}, with the corresponding final t-SNE feature distributions on the CiteSeer dataset further visualized in Figure~\ref{fig:tsne_citeseer}.

Each variant deactivates a specific mechanism corresponding to our core components to validate its necessity and effect on the overall model:
\begin{itemize}
    \item \textbf{w/o $\mathcal{L}_{gcl}$:} Removes the Topology-Aware Graph Contrastive Flow, evaluating its role as a structural false-positive truncation mechanism and its preservation of local structural information under heterophily.
    \item \textbf{w/o $\mathcal{L}_{unsup}$:} Removes the TPR-Guided Unsupervised Alignment Flow, examining the impact of the global semantic bridge on dynamically discovering novel categories and capturing cross-client semantic consistency among unlabeled nodes.
    \item \textbf{w/o TRG: } Removes the TRG module by setting the TPR soft weights ($w_v$) to a uniform value in the unsupervised flow, assessing its mitigation of the neighborhood absorption effect by filtering heterophilous noise during prototype alignment.
\end{itemize}

First, removing the Topology-Aware Graph Contrastive Flow (\textbf{w/o $\mathcal{L}_{gcl}$}) degrades both performance and representation quality. As observed in Figure~\ref{fig:tsne_citeseer}, the absence of $\mathcal{L}_{gcl}$ causes node embeddings to merge into a continuous manifold lacking distinct decision boundaries. Algorithmically, $\mathcal{L}_{gcl}$ introduces a false-positive truncation mechanism that pulls homophilous neighbors together while pushing heterophilous ones apart, forcing representations to respect both semantic boundaries and intrinsic topological structures. Without this guidance, the GNN backbone becomes highly susceptible to neighborhood absorption and over-smoothing. This diminishes intra-class distinctiveness and leads to performance variance, such as the $37.21 \pm 17.91$ variance on Cora demonstrated by the large error bars in Figure~\ref{fig:ablation_bars}.

Second, when the TPR estimation or the unsupervised alignment flow is disabled (\textbf{w/o TRG} or \textbf{w/o $\mathcal{L}_{unsup}$}), Figure~\ref{fig:tsne_citeseer} illustrates that the embedding space becomes fragmented, failing to form compact semantic clusters. The notable performance drop in the \textbf{w/o TRG} variant highlights the necessity of dynamically filtering noisy node-to-prototype assignments. While the TPR metric quantifies node trustworthiness via a logical conjunction of predictive confidence and structural smoothness, disabling TRG forces the unsupervised flow to assign a uniform soft weight to all pseudo-labels. This uniform weighting amplifies unreliable heterophilous signals, exacerbating the representation bias towards inaccurate local prototypes. Similarly, removing the unsupervised flow (\textbf{w/o $\mathcal{L}_{unsup}$}) deprives the unlabeled nodes of the semantic bridge provided by global prototypes, hindering the model's capacity to reliably cluster dynamically discovered novel categories.

Finally, as demonstrated by the Ours variant in Figure~\ref{fig:tsne_citeseer}, the complete GCD-FGL framework integrates these components effectively. By leveraging the TRG mechanism to filter heterophilous noise and dynamically combining the contrastive flow ($\mathcal{L}_{gcl}$) with the unsupervised alignment ($\mathcal{L}_{unsup}$), the Topology-Reliable Semantic Alignment and Discovery process calibrates the isolated node representations. This integration enforces intra-class compactness and inter-class separability, resulting in distinct clustering boundaries even under structural truncation.

\subsection{Hyperparameter Sensitivity}

\begin{figure*}
    \centering
    
    \begin{subfigure}[b]{0.24\textwidth}
        \centering
        \includegraphics[width=\textwidth]{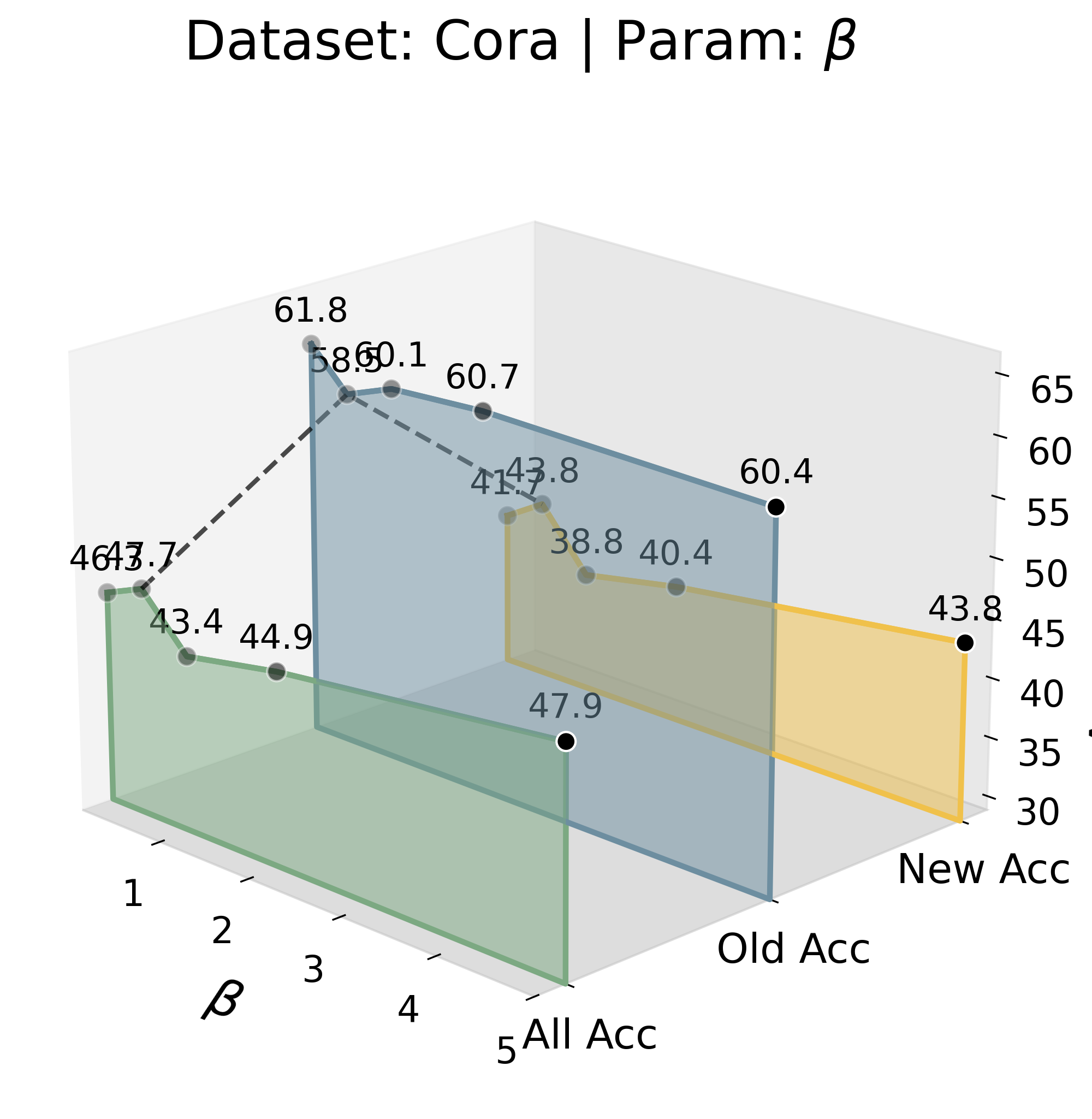}
        \caption{Cora - $\beta$} 
        \label{fig:cora_sgc}
    \end{subfigure}
    \hfill 
    \begin{subfigure}[b]{0.24\textwidth}
        \centering
        \includegraphics[width=\textwidth]{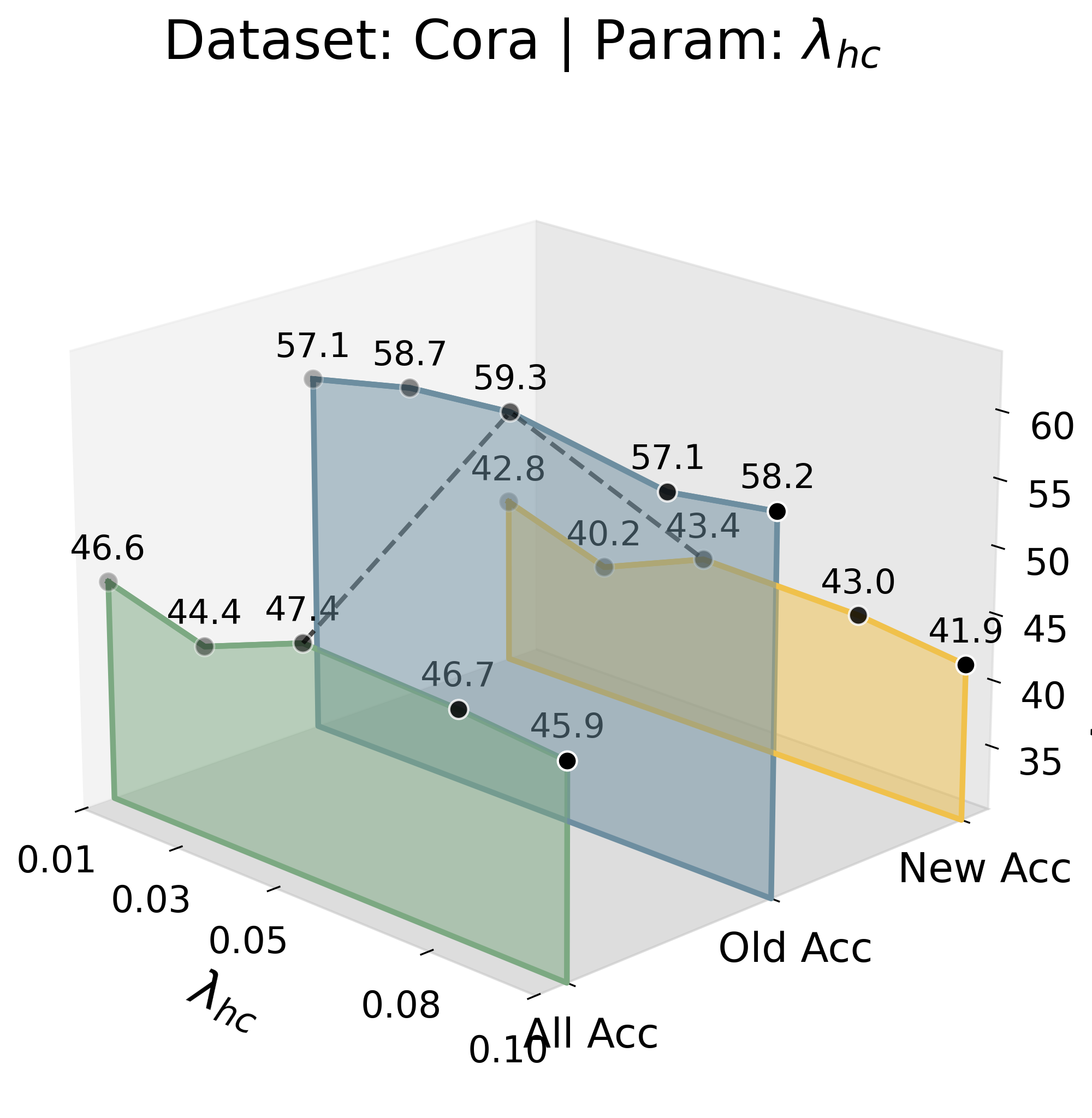}
        \caption{Cora - $\lambda_{hc}$} 
        \label{fig:cora_hc}
    \end{subfigure}
    \hfill
    \begin{subfigure}[b]{0.24\textwidth}
        \centering
        \includegraphics[width=\textwidth]{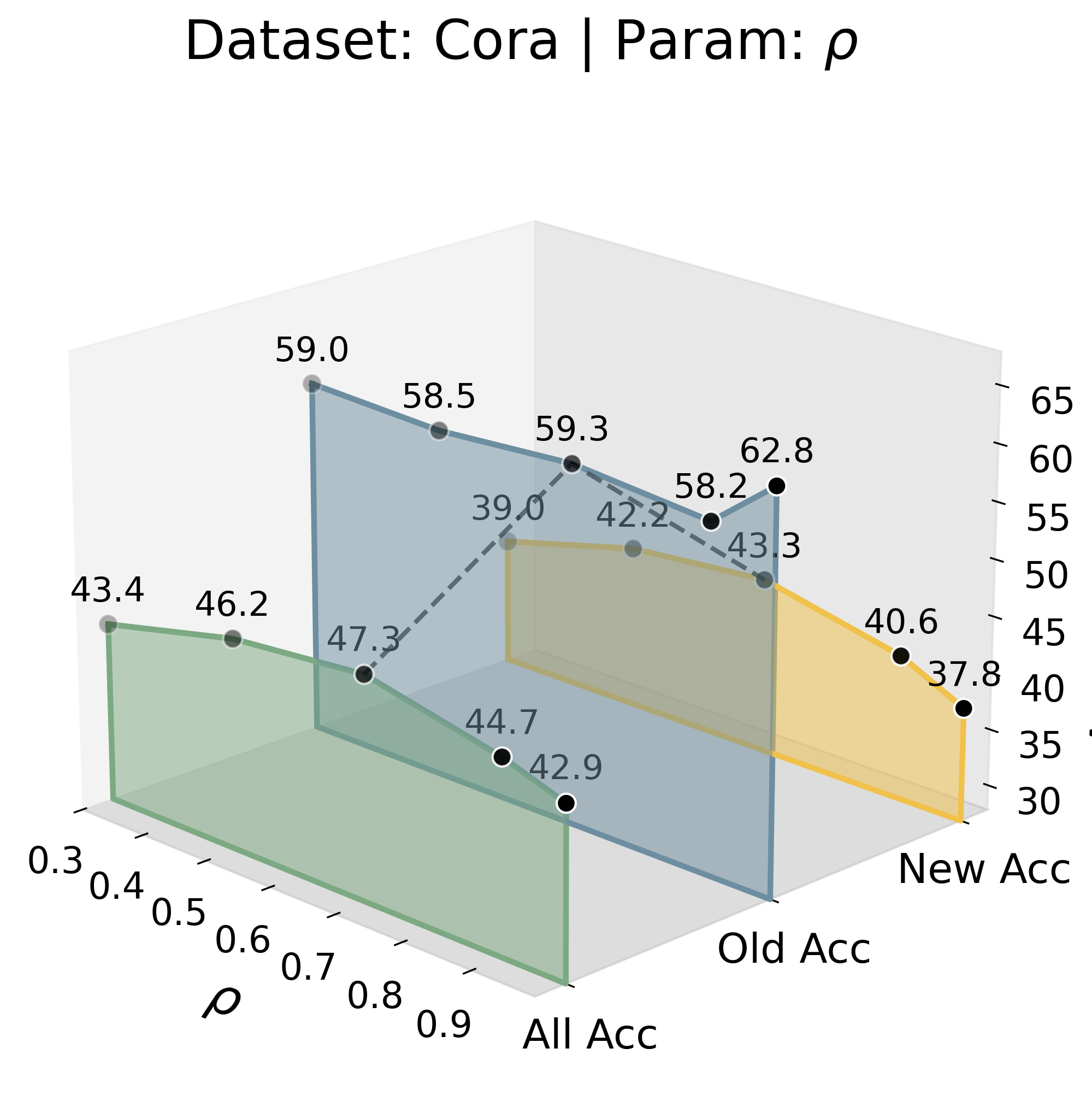}
        \caption{Cora - $\rho$} 
        \label{fig:cora_ema}
    \end{subfigure}
    \hfill
    \begin{subfigure}[b]{0.24\textwidth}
        \centering
        \includegraphics[width=\textwidth]{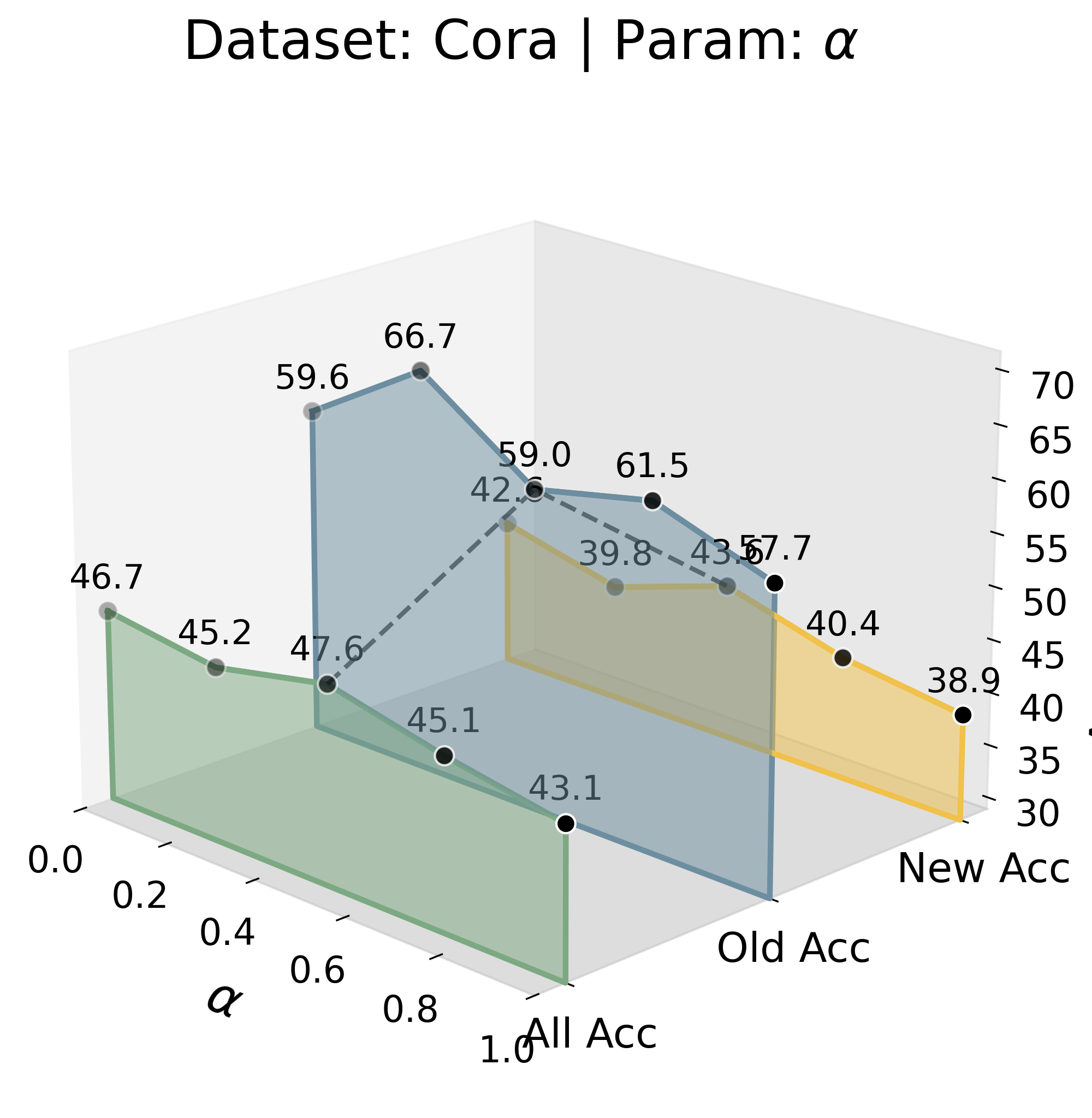}
        \caption{Cora - $\alpha$} 
        \label{fig:cora_pseudo}
    \end{subfigure}
    
    \vspace{0.3cm} 
    
    \begin{subfigure}[b]{0.24\textwidth}
        \centering
        \includegraphics[width=\textwidth]{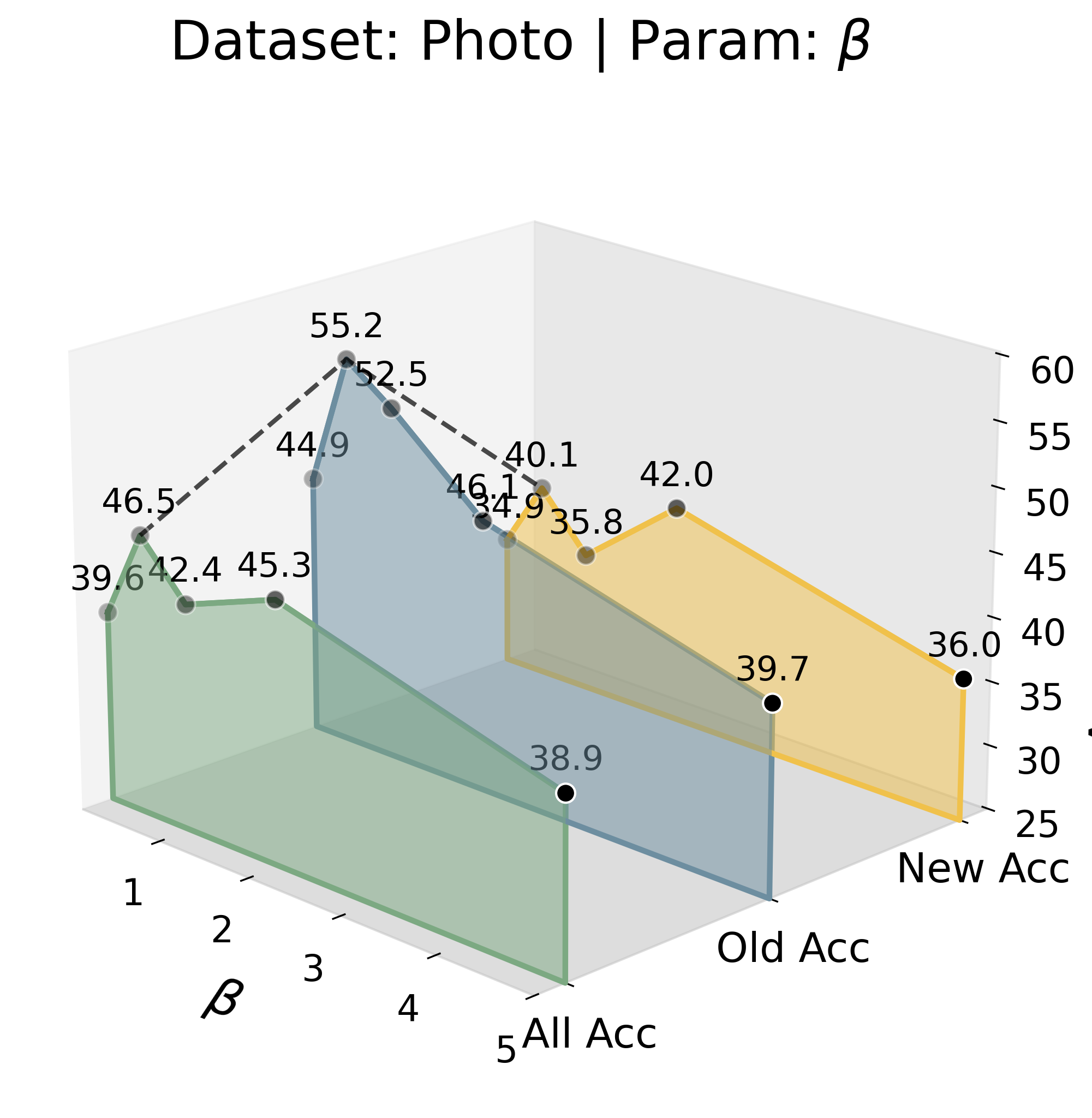}
        \caption{Photo - $\beta$} 
        \label{fig:photo_sgc}
    \end{subfigure}
    \hfill
    \begin{subfigure}[b]{0.24\textwidth}
        \centering
        \includegraphics[width=\textwidth]{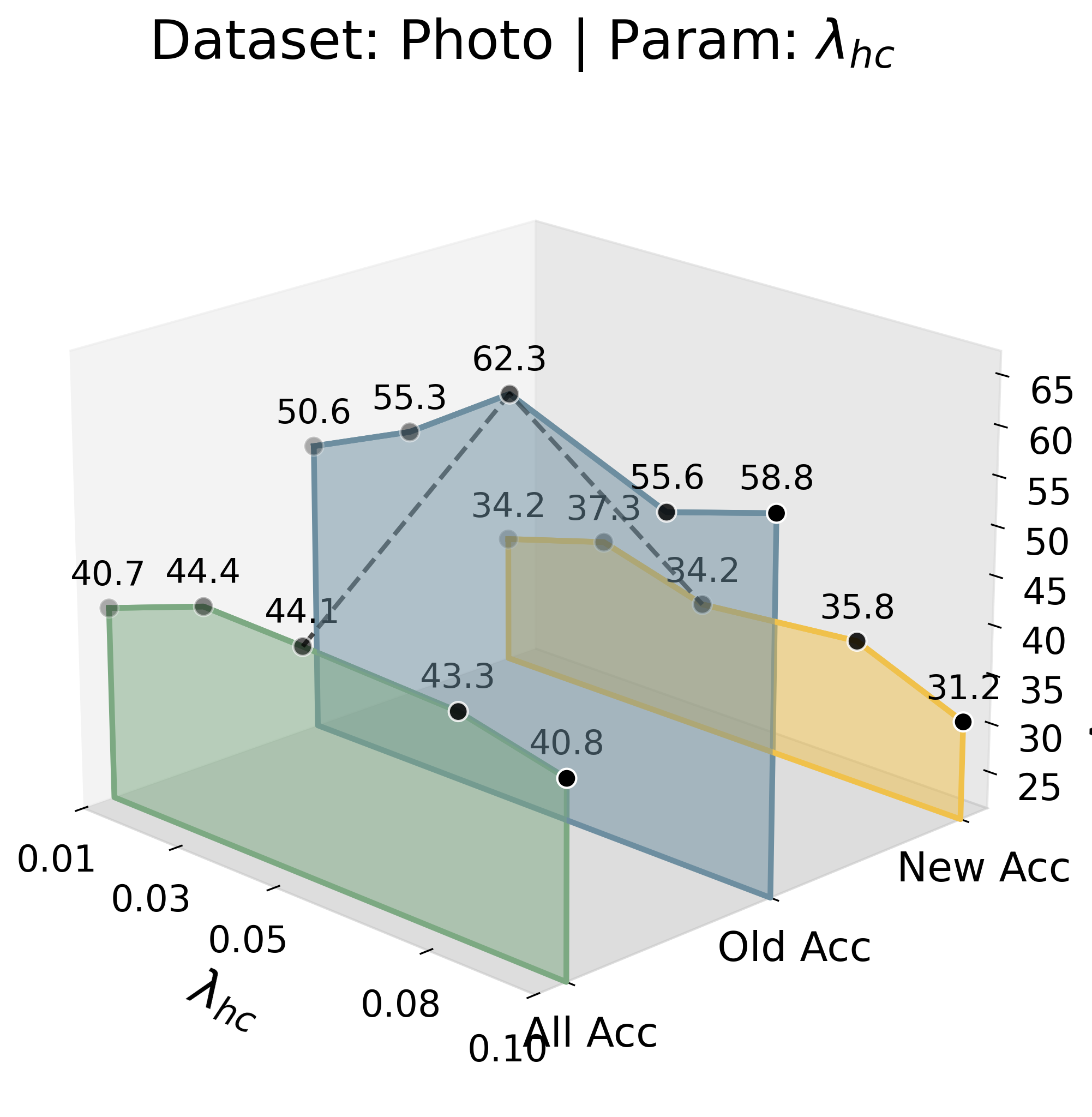}
        \caption{Photo - $\lambda_{hc}$} 
        \label{fig:photo_hc}
    \end{subfigure}
    \hfill
    \begin{subfigure}[b]{0.24\textwidth}
        \centering
        \includegraphics[width=\textwidth]{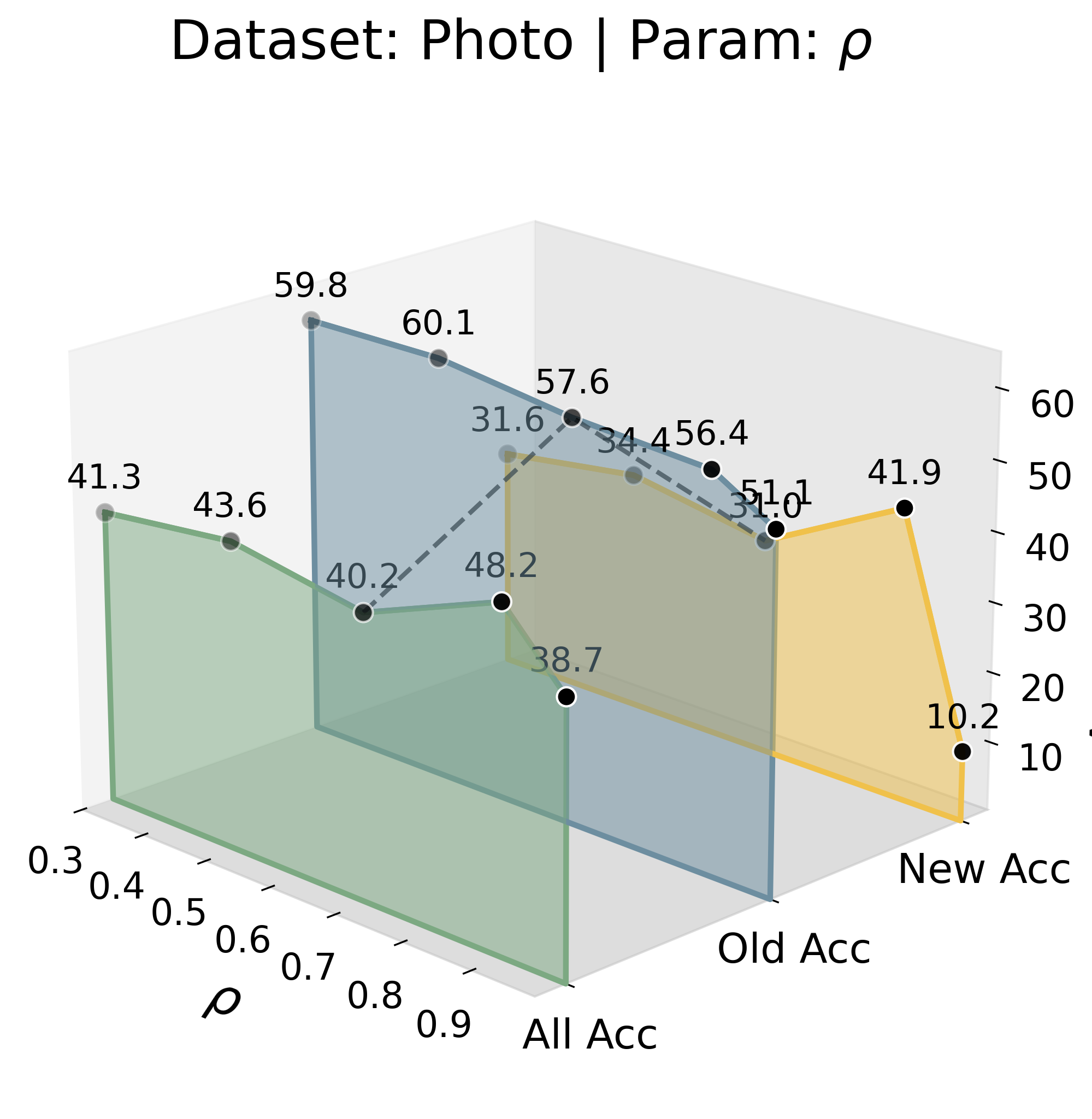}
        \caption{Photo - $\rho$} 
        \label{fig:photo_ema}
    \end{subfigure}
    \hfill
    \begin{subfigure}[b]{0.24\textwidth}
        \centering
        \includegraphics[width=\textwidth]{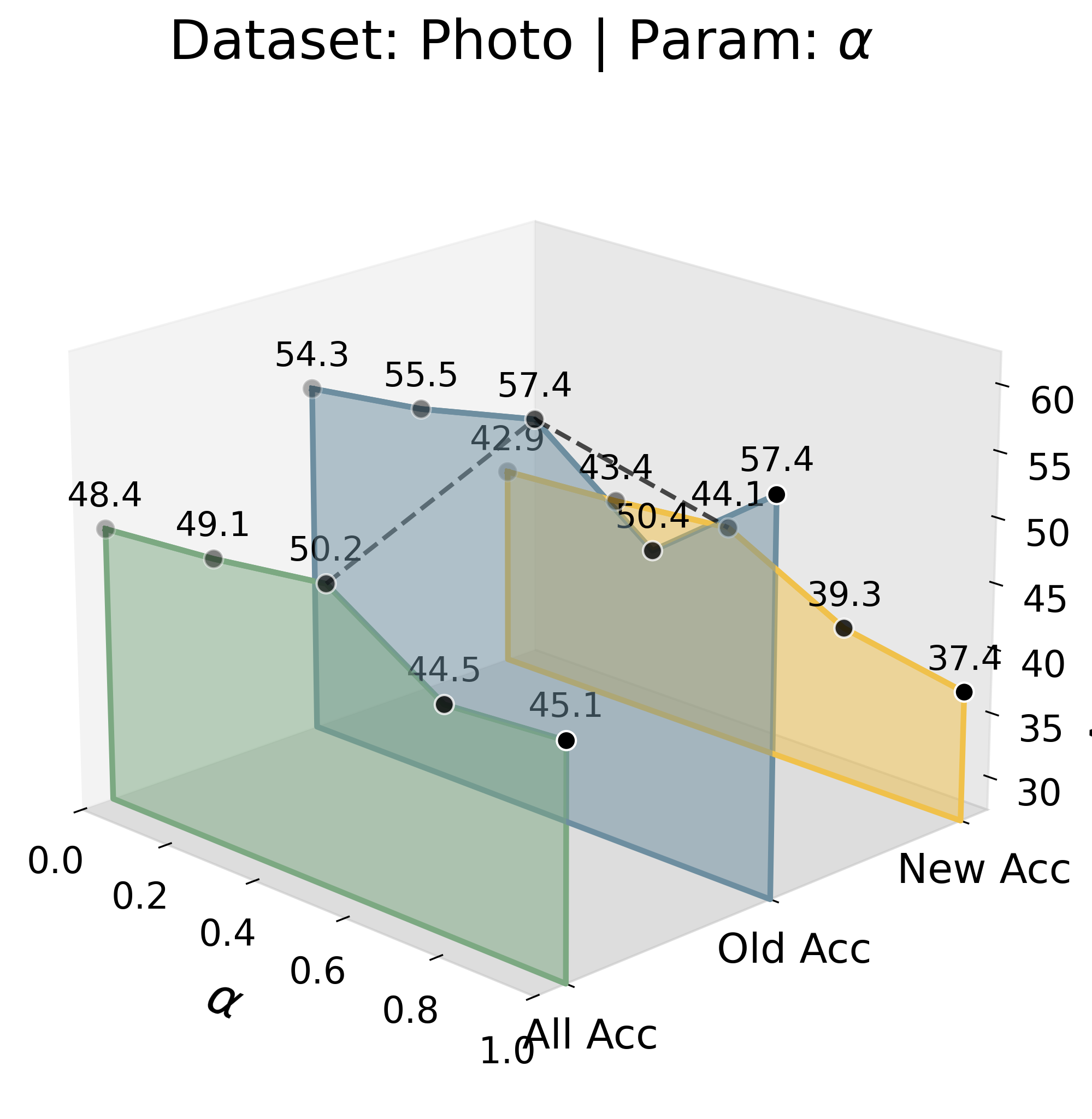}
        \caption{Photo - $\alpha$} 
        \label{fig:photo_pseudo}
    \end{subfigure}
    
    \caption{Hyperparameter sensitivity analysis of core components. We select Cora as a representative small-scale dataset and Photo as a larger-scale dataset. The variables $\beta$, $\lambda_{hc}$, $\rho$, and $\alpha$ denote the contrastive loss weight, hierarchical clustering penalty, EMA momentum, and pseudo-label scaling factor, respectively. The z-axis represents the classification accuracy.}
    \label{fig:hyperparameter_sensitivity}
\end{figure*}

To answer \textbf{Q3}, we systematically analyze the hyperparameter sensitivity of the proposed GCD-FGL framework under the standard distributed setting (partitioned via the Louvain algorithm). We select Cora and Amazon Photo as representative small-scale and large-scale datasets, respectively. Figure~\ref{fig:hyperparameter_sensitivity} visualizes the resulting performance variations across varying values for four key parameters: the Topology-Aware Graph Contrastive Flow weight ($\beta$), the Constrained Hierarchical Clustering penalty ($\lambda_{hc}$), the EMA Momentum Update decay factor ($\rho$), and the dynamic threshold scaling factor ($\alpha$). The default values for each parameter are marked with dashed lines in the respective subplots.

Empirical results demonstrate stable performance across most evaluated parameter ranges. Specifically, parameters governing temporal prototype smoothing ($\rho$), server-side cluster regularization ($\lambda_{hc}$), and confident pseudo-label filtering ($\alpha$) exhibit minimal performance fluctuations on both datasets. This indicates that the dynamic thresholding and hierarchical routing mechanisms are stable, reducing the reliance on precise manual tuning.

Sensitivity is observed primarily regarding the contrastive loss weight ($\beta$) on the denser Amazon Photo dataset (Figure~\ref{fig:photo_sgc}). In graphs with dense homophilous connections, such as Photo, the structural contrastive flow ($\mathcal{L}_{gcl}$) has a direct impact on the embedding space. Setting $\beta$ too low fails to fully exploit local topological homophily to counteract the neighborhood absorption effect. Conversely, a high $\beta$ value risks over-regularizing the representation space, potentially interfering with the TPR-Guided Unsupervised Flow and affecting the convergence of global prototypes. 

In contrast, on the sparser Cora dataset (Figure~\ref{fig:cora_sgc}), the model is less sensitive to $\beta$ variations due to its lower structural density. Ultimately, despite the localized sensitivity on denser graphs, the overall performance variations remain bounded, suggesting that GCD-FGL achieves reliable generalization without relying on extensive parameter tuning.

\subsection{Robustness}

\begin{table*}
\centering
\caption{Robustness results of GCD-FGL under random label sparsity across five graph datasets. The best results are highlighted in bold, and the second-best results are underlined. The `Gain` column shows the improvement between GCD-FGL and the best baseline.}
\label{tab:robustness_random_label_sparsity}
\resizebox{\textwidth}{!}{
\begin{tabular}{l|l|ccccccc|c}
\toprule
Datasets & Metrics & AutoNovel & ORCA & GCD & SimGCD & FedGCD & SWIRL & GCD-FGL & Gain \\
\midrule
\multirow{4}{*}{Cora} & \cellcolor{gray!10}HRScore & \cellcolor{gray!10}$34.12_{{\scriptstyle\,\pm\,1.77}}$ & \cellcolor{gray!10}$23.39_{{\scriptstyle\,\pm\,0.65}}$ & \cellcolor{gray!10}$32.67_{{\scriptstyle\,\pm\,2.97}}$ & \cellcolor{gray!10}$19.15_{{\scriptstyle\,\pm\,2.68}}$ & \cellcolor{gray!10}$36.38_{{\scriptstyle\,\pm\,8.25}}$ & \cellcolor{gray!10}$\underline{43.26}_{{\scriptstyle\,\pm\,1.70}}$ & \cellcolor{gray!10}$\mathbf{55.77_{{\scriptstyle\,\pm\,2.25}}}$ & \cellcolor{gray!10}+12.51 \\
\cmidrule{2-10}
 & Old  & $42.05_{{\scriptstyle\,\pm\,5.16}}$ & $39.59_{{\scriptstyle\,\pm\,0.35}}$ & $32.31_{{\scriptstyle\,\pm\,5.79}}$ & $37.99_{{\scriptstyle\,\pm\,1.43}}$ & $37.95_{{\scriptstyle\,\pm\,17.40}}$ & $\underline{47.63}_{{\scriptstyle\,\pm\,3.20}}$ & $\mathbf{62.82_{{\scriptstyle\,\pm\,3.10}}}$ & +15.19 \\
 & New  & $28.71_{{\scriptstyle\,\pm\,0.71}}$ & $16.60_{{\scriptstyle\,\pm\,0.31}}$ & $33.03_{{\scriptstyle\,\pm\,0.45}}$ & $12.80_{{\scriptstyle\,\pm\,0.30}}$ & $34.93_{{\scriptstyle\,\pm\,3.79}}$ & $\underline{39.62}_{{\scriptstyle\,\pm\,1.80}}$ & $\mathbf{50.15_{{\scriptstyle\,\pm\,2.85}}}$ & +10.53 \\
 & All  & $30.55_{{\scriptstyle\,\pm\,0.10}}$ & $34.38_{{\scriptstyle\,\pm\,0.54}}$ & $32.93_{{\scriptstyle\,\pm\,0.74}}$ & $32.95_{{\scriptstyle\,\pm\,0.23}}$ & $35.34_{{\scriptstyle\,\pm\,4.61}}$ & $\underline{40.72}_{{\scriptstyle\,\pm\,1.30}}$ & $\mathbf{51.89_{{\scriptstyle\,\pm\,2.16}}}$ & +11.17 \\
\midrule
\multirow{4}{*}{CiteSeer} & \cellcolor{gray!10}HRScore & \cellcolor{gray!10}$42.32_{{\scriptstyle\,\pm\,0.45}}$ & \cellcolor{gray!10}$47.82_{{\scriptstyle\,\pm\,0.38}}$ & \cellcolor{gray!10}$37.35_{{\scriptstyle\,\pm\,1.35}}$ & \cellcolor{gray!10}$40.60_{{\scriptstyle\,\pm\,0.65}}$ & \cellcolor{gray!10}$29.23_{{\scriptstyle\,\pm\,1.63}}$ & \cellcolor{gray!10}$\underline{54.49}_{{\scriptstyle\,\pm\,7.33}}$ & \cellcolor{gray!10}$\mathbf{61.15_{{\scriptstyle\,\pm\,2.17}}}$ & \cellcolor{gray!10}+6.66 \\
\cmidrule{2-10}
 & Old  & $47.22_{{\scriptstyle\,\pm\,0.84}}$ & $\underline{66.07}_{{\scriptstyle\,\pm\,0.63}}$ & $36.05_{{\scriptstyle\,\pm\,1.78}}$ & $38.06_{{\scriptstyle\,\pm\,0.67}}$ & $61.84_{{\scriptstyle\,\pm\,0.66}}$ & $51.35_{{\scriptstyle\,\pm\,12.85}}$ & $\mathbf{69.59_{{\scriptstyle\,\pm\,4.64}}}$ & +3.52 \\
 & New  & $38.34_{{\scriptstyle\,\pm\,0.49}}$ & $37.47_{{\scriptstyle\,\pm\,0.42}}$ & $38.75_{{\scriptstyle\,\pm\,2.06}}$ & $43.51_{{\scriptstyle\,\pm\,1.20}}$ & $18.50_{{\scriptstyle\,\pm\,1.26}}$ & $\mathbf{58.05_{{\scriptstyle\,\pm\,2.68}}}$ & $\underline{54.51}_{{\scriptstyle\,\pm\,3.81}}$ & -3.54 \\
 & All  & $40.30_{{\scriptstyle\,\pm\,0.39}}$ & $43.80_{{\scriptstyle\,\pm\,0.66}}$ & $38.16_{{\scriptstyle\,\pm\,1.56}}$ & $42.31_{{\scriptstyle\,\pm\,0.85}}$ & $29.81_{{\scriptstyle\,\pm\,1.65}}$ & $\underline{56.57}_{{\scriptstyle\,\pm\,1.81}}$ & $\mathbf{60.21_{{\scriptstyle\,\pm\,0.59}}}$ & +3.64 \\
\midrule
\multirow{4}{*}{CS} & \cellcolor{gray!10}HRScore & \cellcolor{gray!10}$\underline{58.09}_{{\scriptstyle\,\pm\,6.97}}$ & \cellcolor{gray!10}$20.63_{{\scriptstyle\,\pm\,0.26}}$ & \cellcolor{gray!10}$55.84_{{\scriptstyle\,\pm\,2.11}}$ & \cellcolor{gray!10}$54.08_{{\scriptstyle\,\pm\,0.09}}$ & \cellcolor{gray!10}$25.35_{{\scriptstyle\,\pm\,6.02}}$ & \cellcolor{gray!10}$56.62_{{\scriptstyle\,\pm\,8.98}}$ & \cellcolor{gray!10}$\mathbf{69.32_{{\scriptstyle\,\pm\,16.74}}}$ & \cellcolor{gray!10}+11.23 \\
\cmidrule{2-10}
 & Old  & $59.60_{{\scriptstyle\,\pm\,14.23}}$ & $61.17_{{\scriptstyle\,\pm\,0.19}}$ & $70.51_{{\scriptstyle\,\pm\,5.01}}$ & $56.19_{{\scriptstyle\,\pm\,0.12}}$ & $\underline{77.78}_{{\scriptstyle\,\pm\,4.57}}$ & $52.72_{{\scriptstyle\,\pm\,15.50}}$ & $\mathbf{79.06_{{\scriptstyle\,\pm\,28.59}}}$ & +1.28 \\
 & New  & $56.66_{{\scriptstyle\,\pm\,3.24}}$ & $12.41_{{\scriptstyle\,\pm\,0.03}}$ & $46.23_{{\scriptstyle\,\pm\,1.92}}$ & $52.13_{{\scriptstyle\,\pm\,0.12}}$ & $15.14_{{\scriptstyle\,\pm\,4.29}}$ & $\underline{61.14}_{{\scriptstyle\,\pm\,1.95}}$ & $\mathbf{61.72_{{\scriptstyle\,\pm\,4.34}}}$ & +0.58 \\
 & All  & $57.06_{{\scriptstyle\,\pm\,1.08}}$ & $54.41_{{\scriptstyle\,\pm\,0.02}}$ & $49.59_{{\scriptstyle\,\pm\,1.47}}$ & $52.69_{{\scriptstyle\,\pm\,0.11}}$ & $23.82_{{\scriptstyle\,\pm\,3.84}}$ & $\underline{59.98}_{{\scriptstyle\,\pm\,1.79}}$ & $\mathbf{64.12_{{\scriptstyle\,\pm\,0.34}}}$ & +4.14 \\
\midrule
\multirow{4}{*}{Photo} & \cellcolor{gray!10}HRScore & \cellcolor{gray!10}$41.65_{{\scriptstyle\,\pm\,5.29}}$ & \cellcolor{gray!10}$36.59_{{\scriptstyle\,\pm\,5.71}}$ & \cellcolor{gray!10}$42.13_{{\scriptstyle\,\pm\,2.58}}$ & \cellcolor{gray!10}$31.56_{{\scriptstyle\,\pm\,0.72}}$ & \cellcolor{gray!10}$43.57_{{\scriptstyle\,\pm\,9.16}}$ & \cellcolor{gray!10}$\underline{45.34}_{{\scriptstyle\,\pm\,2.96}}$ & \cellcolor{gray!10}$\mathbf{48.54_{{\scriptstyle\,\pm\,19.12}}}$ & \cellcolor{gray!10}+3.20 \\
\cmidrule{2-10}
 & Old  & $40.60_{{\scriptstyle\,\pm\,5.45}}$ & $30.48_{{\scriptstyle\,\pm\,7.61}}$ & $36.84_{{\scriptstyle\,\pm\,3.74}}$ & $24.98_{{\scriptstyle\,\pm\,0.90}}$ & $45.40_{{\scriptstyle\,\pm\,14.63}}$ & $\underline{54.03}_{{\scriptstyle\,\pm\,6.58}}$ & $\mathbf{60.15_{{\scriptstyle\,\pm\,10.97}}}$ & +6.12 \\
 & New  & $42.75_{{\scriptstyle\,\pm\,9.36}}$ & $\underline{45.78}_{{\scriptstyle\,\pm\,4.98}}$ & $\mathbf{49.18_{{\scriptstyle\,\pm\,2.24}}}$ & $42.84_{{\scriptstyle\,\pm\,0.33}}$ & $41.89_{{\scriptstyle\,\pm\,5.55}}$ & $39.04_{{\scriptstyle\,\pm\,2.74}}$ & $40.69_{{\scriptstyle\,\pm\,26.39}}$ & -8.49 \\
 & All  & $42.12_{{\scriptstyle\,\pm\,5.70}}$ & $41.26_{{\scriptstyle\,\pm\,2.28}}$ & $\underline{45.53}_{{\scriptstyle\,\pm\,1.61}}$ & $37.56_{{\scriptstyle\,\pm\,0.24}}$ & $42.93_{{\scriptstyle\,\pm\,0.82}}$ & $43.47_{{\scriptstyle\,\pm\,1.12}}$ & $\mathbf{54.40_{{\scriptstyle\,\pm\,7.65}}}$ & +8.87 \\
\midrule
\multirow{4}{*}{Computers} & \cellcolor{gray!10}HRScore & \cellcolor{gray!10}$39.48_{{\scriptstyle\,\pm\,5.49}}$ & \cellcolor{gray!10}$46.64_{{\scriptstyle\,\pm\,7.60}}$ & \cellcolor{gray!10}$44.85_{{\scriptstyle\,\pm\,3.63}}$ & \cellcolor{gray!10}$\underline{48.72}_{{\scriptstyle\,\pm\,4.73}}$ & \cellcolor{gray!10}$47.80_{{\scriptstyle\,\pm\,6.53}}$ & \cellcolor{gray!10}$39.18_{{\scriptstyle\,\pm\,6.38}}$ & \cellcolor{gray!10}$\mathbf{49.23_{{\scriptstyle\,\pm\,0.54}}}$ & \cellcolor{gray!10}+0.51 \\
\cmidrule{2-10}
 & Old  & $48.96_{{\scriptstyle\,\pm\,10.59}}$ & $\underline{51.85}_{{\scriptstyle\,\pm\,5.55}}$ & $41.65_{{\scriptstyle\,\pm\,6.03}}$ & $51.21_{{\scriptstyle\,\pm\,4.05}}$ & $47.60_{{\scriptstyle\,\pm\,8.99}}$ & $35.47_{{\scriptstyle\,\pm\,8.95}}$ & $\mathbf{59.46_{{\scriptstyle\,\pm\,0.66}}}$ & +7.61 \\
 & New  & $33.08_{{\scriptstyle\,\pm\,6.01}}$ & $42.37_{{\scriptstyle\,\pm\,11.99}}$ & $\mathbf{48.55_{{\scriptstyle\,\pm\,2.33}}}$ & $46.48_{{\scriptstyle\,\pm\,7.15}}$ & $\underline{48.00}_{{\scriptstyle\,\pm\,9.48}}$ & $43.72_{{\scriptstyle\,\pm\,8.25}}$ & $42.01_{{\scriptstyle\,\pm\,0.72}}$ & -6.54 \\
 & All  & $41.32_{{\scriptstyle\,\pm\,6.77}}$ & $47.29_{{\scriptstyle\,\pm\,3.97}}$ & $44.97_{{\scriptstyle\,\pm\,2.53}}$ & $\underline{48.93}_{{\scriptstyle\,\pm\,2.10}}$ & $47.80_{{\scriptstyle\,\pm\,3.52}}$ & $39.44_{{\scriptstyle\,\pm\,1.30}}$ & $\mathbf{51.54_{{\scriptstyle\,\pm\,0.39}}}$ & +2.61 \\
\bottomrule
\multicolumn{10}{l}{\footnotesize \textit{Note:} The random label sparsity rate is set to 10\%.} \\
\end{tabular}
}
\end{table*}

To answer \textbf{Q4}, we evaluate the robustness of GCD-FGL under label sparsity within the distributed federated setting (partitioned via the Louvain algorithm). We set the random label sparsity rate to 10\% across all clients. In FGGCD, limited supervisory signals, compounded by heterogeneous and partially overlapping label spaces, can predispose models to overfit local known categories, thereby limiting their capacity to discover novel categories. Quantitative results under this sparse regime are detailed in Table~\ref{tab:robustness_random_label_sparsity}.

Under these conditions, GCD-FGL demonstrates stable performance, achieving the highest HRScore and overall accuracy (All Acc) on all five benchmark datasets (Cora, CiteSeer, Coauthor CS, Amazon Photo, and Amazon Computers). For instance, on the CS dataset, the framework yields an HRScore of 69.32\% and a novel category accuracy (New Acc) of 61.72\%. On the Computers dataset, it records an HRScore of 49.23\%. This performance is primarily attributed to the integration of the TRG mechanism and the client-side semantic alignment flow. When labeled anchors are scarce, these components filter heterophilous noise and regularize node-to-prototype assignments based on structural homophily, enabling the model to leverage unlabeled nodes for representation learning.

While certain baseline methods perform well on isolated metrics or datasets, such as GCD regarding novel category discovery (New Acc) on the Amazon Photo dataset or ORCA regarding known category retention (Old Acc) on CiteSeer, they exhibit performance imbalances under sparse conditions. For example, on the CS dataset, FedGCD achieves an Old Acc of 77.78\% but a degraded New Acc of 15.14\%. This indicates that, without adequate structural regularization, baselines tend to exhibit local representation bias, overfit to the limited known labels, and struggle to generalize to novel categories. In contrast, GCD-FGL mitigates semantic divergence, maintaining balanced predictions across evaluation metrics despite sparsity.

\subsection{Efficiency}

\begin{figure}
    \centering
    \begin{subfigure}[b]{0.48\columnwidth}
        \centering
        \includegraphics[width=\linewidth]{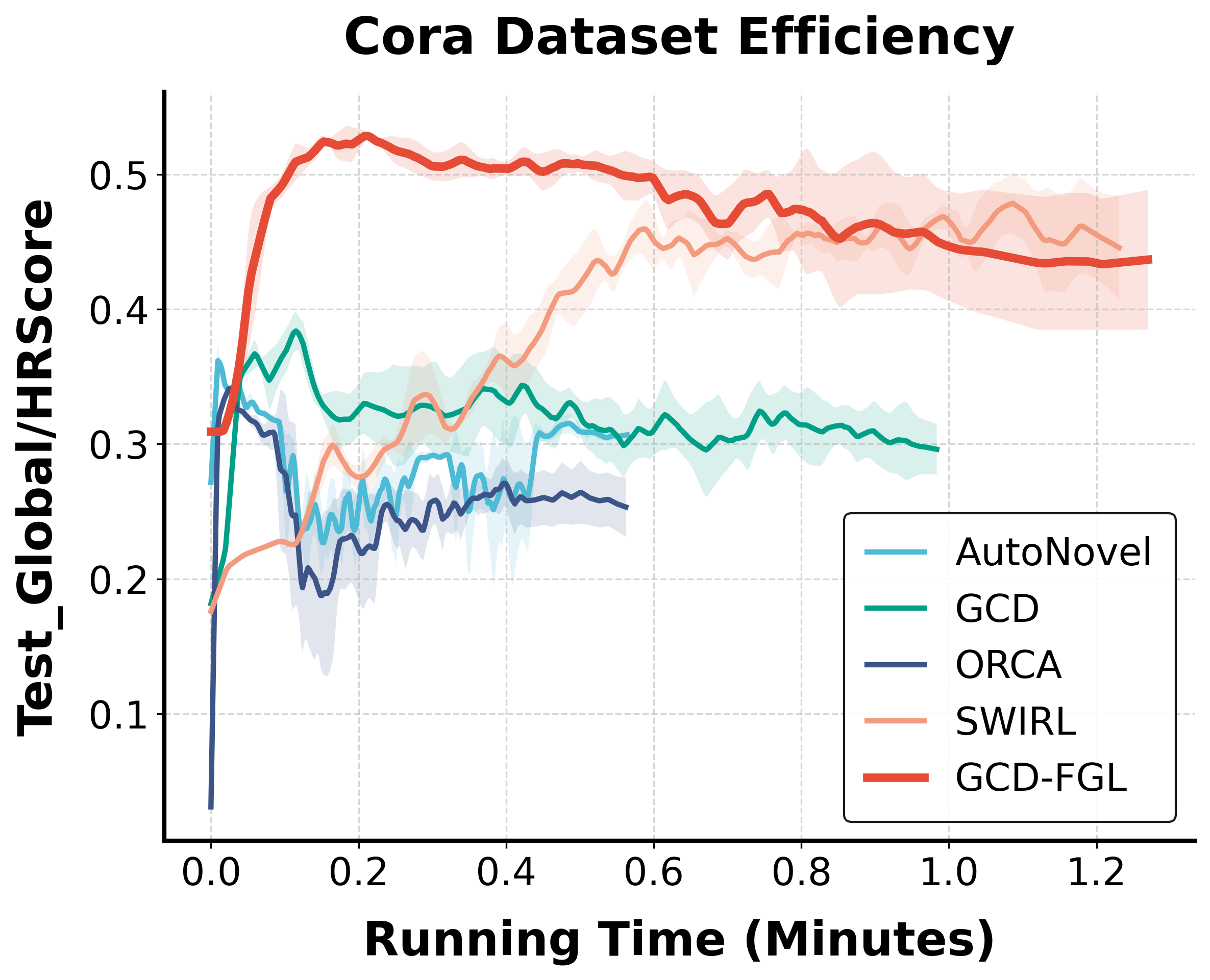}
        \caption{Cora}
        \label{fig:eff_cora}
    \end{subfigure}
    \hfill 
    \begin{subfigure}[b]{0.48\columnwidth}
        \centering
        \includegraphics[width=\linewidth]{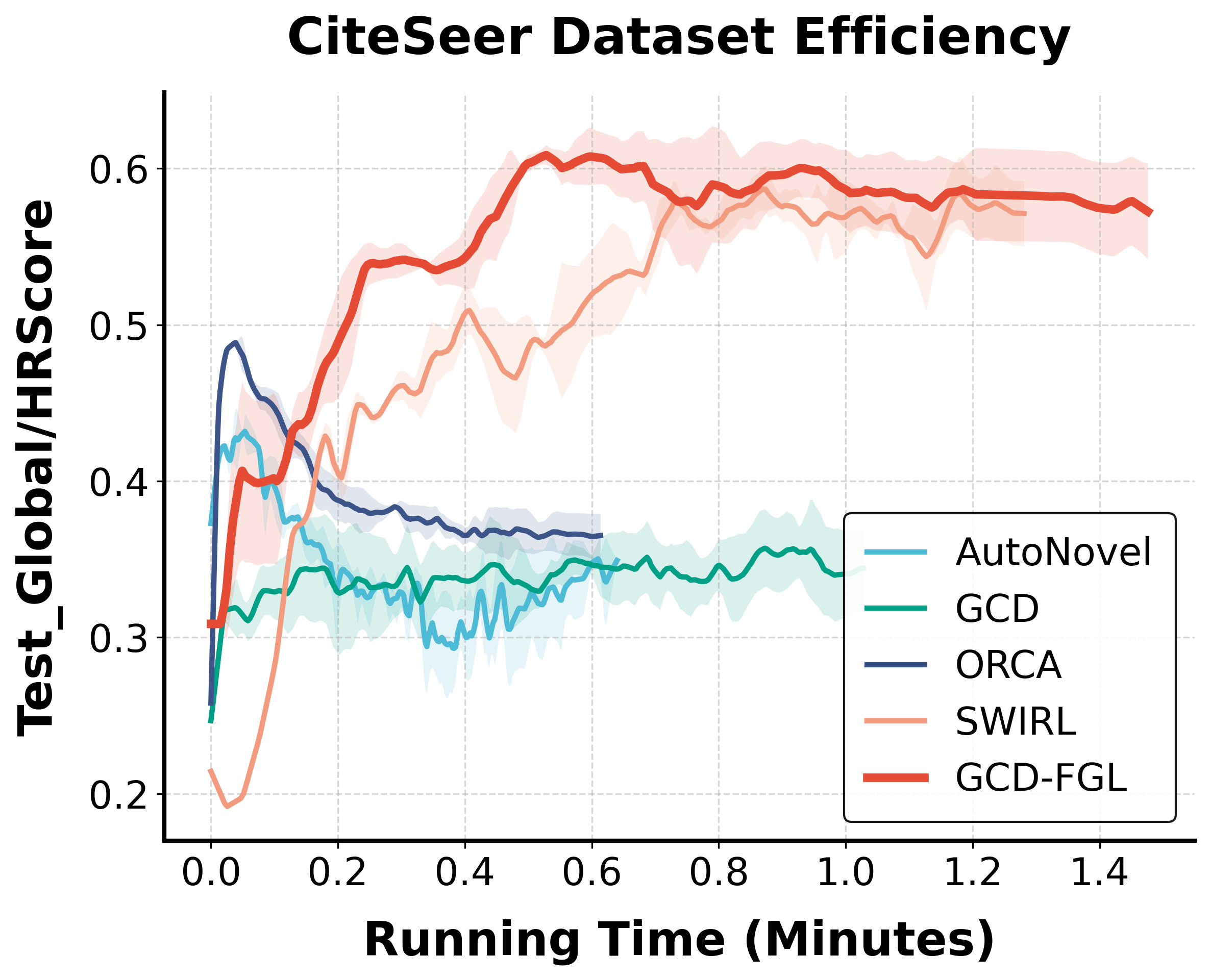}
        \caption{CiteSeer}
        \label{fig:eff_citeseer}
    \end{subfigure}
    
    
    \begin{subfigure}[b]{0.48\columnwidth}
        \centering
        \includegraphics[width=\linewidth]{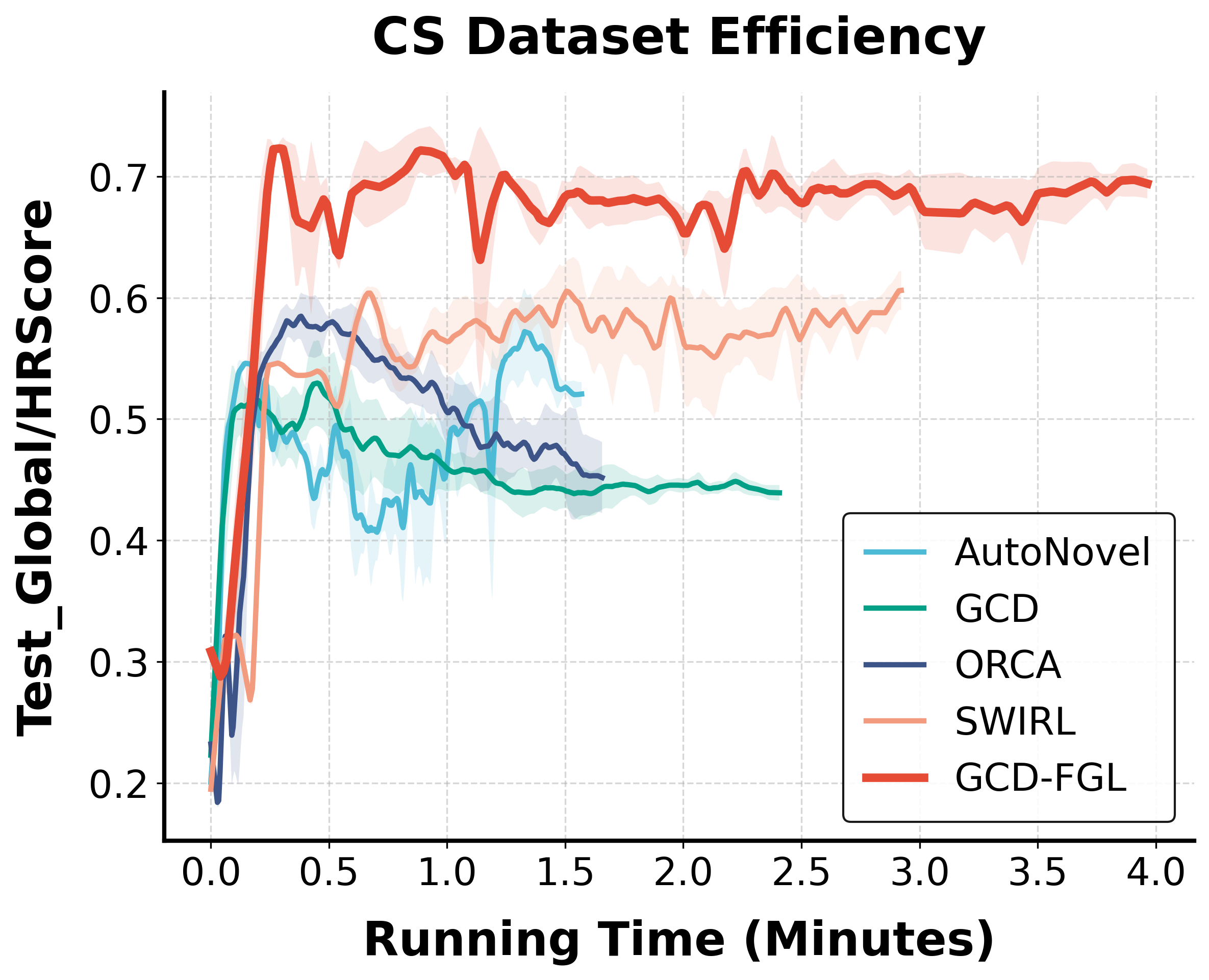}
        \caption{Coauthor CS} 
        \label{fig:eff_cs}
    \end{subfigure}
    \hfill
    \begin{subfigure}[b]{0.48\columnwidth}
        \centering
        \includegraphics[width=\linewidth]{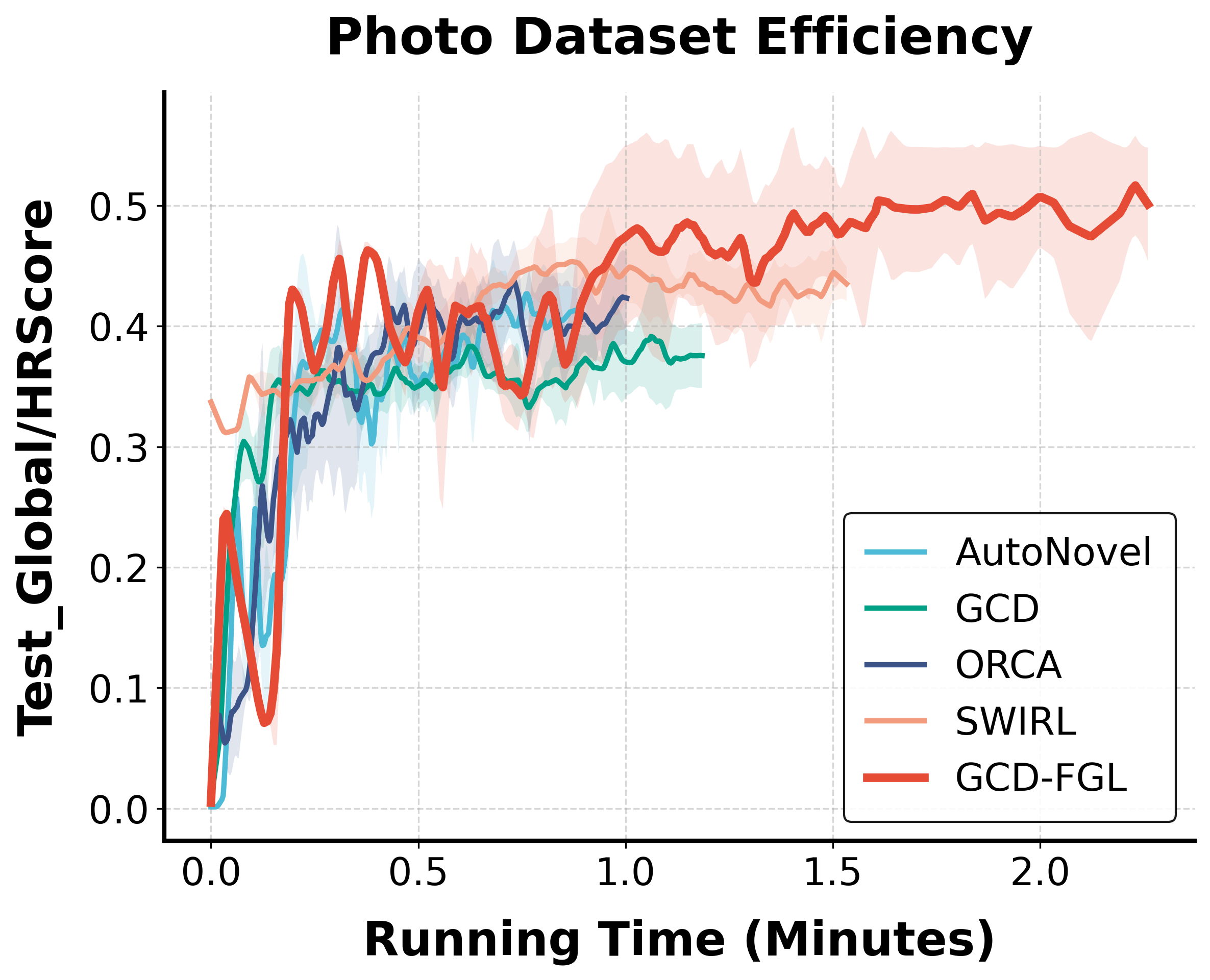}
        \caption{Amazon Photo} 
        \label{fig:eff_photo}
    \end{subfigure}
    
    \caption{Training efficiency on four datasets. Solid lines denote mean performance over multiple runs, and shaded regions indicate standard deviation.}
    \label{fig:efficiency_all}
\end{figure}

To answer \textbf{Q5}, we evaluate the training efficiency of GCD-FGL by analyzing the evolution of predictive performance relative to wall-clock running time. This evaluation is conducted under the distributed setting (partitioned via the Louvain algorithm). Figure~\ref{fig:efficiency_all} illustrates the training efficiency curves, reporting the mean performance and standard deviation across four datasets (Cora, CiteSeer, Coauthor CS, and Amazon Photo).

Regarding learning dynamics, the performance curves of GCD-FGL exhibit faster initial convergence across all datasets, reaching performance plateaus earlier than the baselines. This efficiency is driven by the Density-Aware Aggregation and the Hierarchical Prototype Alignment mechanisms. By aligning the global semantic space and filtering out unreliable local updates, these components provide consistent optimization directions, reducing the total number of communication rounds required for convergence. However, a late-stage oscillation is observed on the Cora dataset, where classification accuracy degrades slightly during the final training epochs, likely due to over-regularization on its relatively sparse topology.

\begin{figure}
    \centering
    \includegraphics[width=1\linewidth]{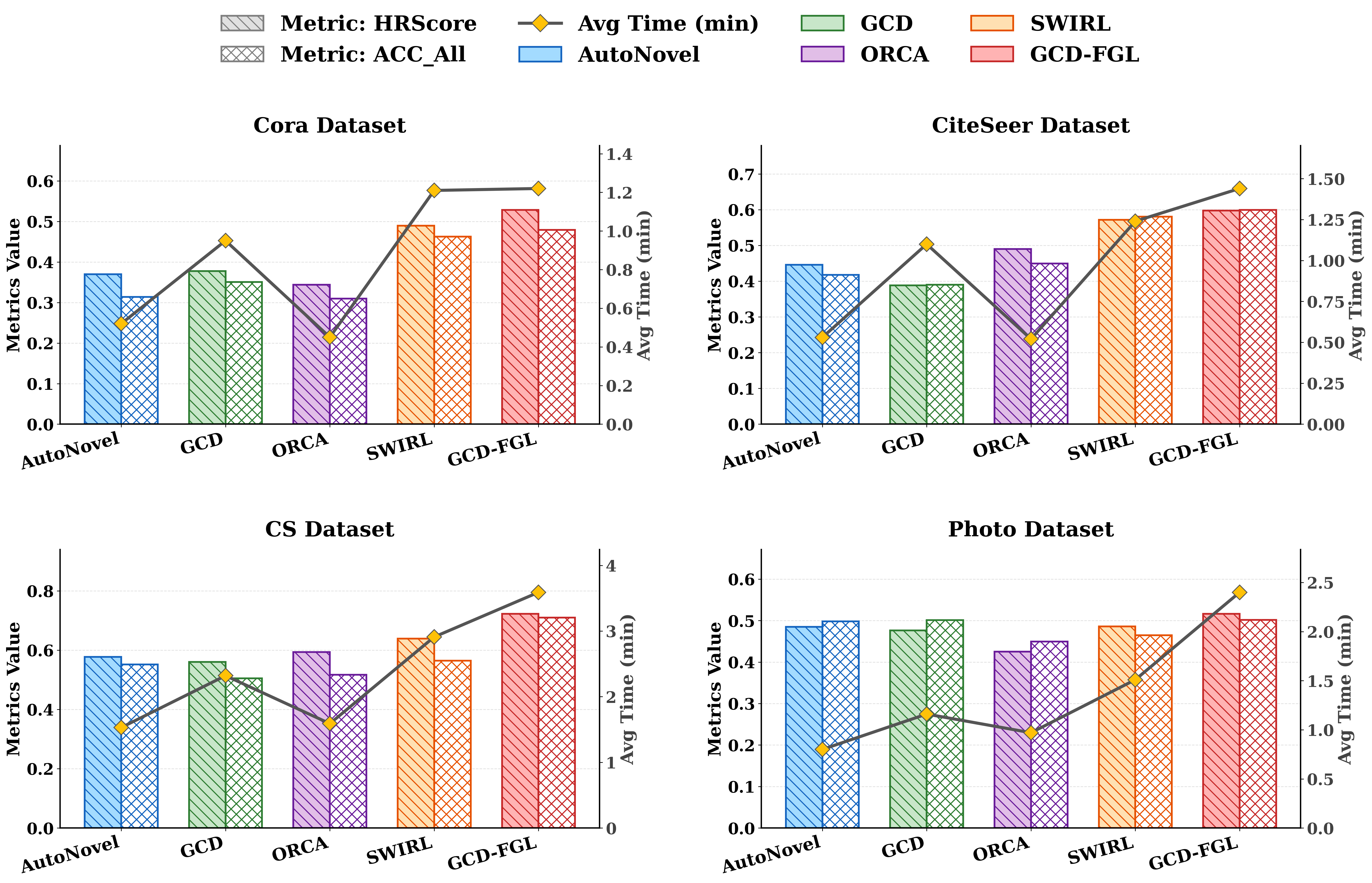}
    \caption{Comparison of HRScore, All Acc, and average running time across four datasets.}
\label{fig:efficiency_comparison}
\end{figure}

In terms of computational cost, the efficiency of GCD-FGL shows a dependence on graph scale, as further visualized in Figure~\ref{fig:efficiency_comparison}. On smaller datasets (Cora and CiteSeer), the total running time is comparable to the baselines. Conversely, on denser and larger datasets (CS and Photo), GCD-FGL incurs an increase in computational overhead per communication round. This discrepancy stems from the TRG mechanism and the Topology-Aware Graph Contrastive Flow calculations. As the graph volume expands, computing local structural smoothness and executing the false-positive truncation mechanism introduce additional computational cost.

Among the baselines, SWIRL demonstrates competitive convergence stability. As an architecture tailored for holistic GGCD, it leverages topological dependencies better than image-data GCD baselines (AutoNovel, ORCA, and GCD), which exhibit performance bottlenecks on these datasets. Nevertheless, lacking dedicated mechanisms to mitigate subgraph distribution discrepancies and heterogeneous label spaces, SWIRL's convergence trajectory ultimately plateaus below that of GCD-FGL. Overall, while GCD-FGL incurs an increased per-round computational cost on large-scale graphs, Figure~\ref{fig:efficiency_comparison} demonstrates that this overhead is offset by its faster convergence rate and higher final accuracy, presenting a practical trade-off for realistic FGGCD scenarios.

\section{Conclusion}
\label{sec:conclusion}

In this paper, we propose GCD-FGL, a framework tailored for FGGCD. A primary challenge in FGGCD is the consistent alignment of dynamically discovered novel categories across isolated and heterogeneous clients, which complicates effective global knowledge aggregation. To address this issue, our method integrates a client-side Topology-Reliable Semantic Alignment and Discovery process, guided by the TRG mechanism, with a server-side Hierarchical Prototype Alignment strategy. Through this architecture, the proposed framework achieves cross-client alignment for novel categories and establishes distinct feature spaces that respect both semantic boundaries and topological homophily.

Experiments across five benchmark datasets demonstrate that GCD-FGL achieves state-of-the-art performance. Robustness and ablation analyses further validate the stability of our method, demonstrating its resilience to label sparsity without exacerbating catastrophic forgetting. While GCD-FGL improves accuracy and stability, it incurs a trade-off in training efficiency. The local Topology-Aware Graph Contrastive Flow and the global Hierarchical Prototype Alignment procedures entail higher computational overhead and longer per-round running times compared to baseline methods. Nevertheless, by addressing the cross-client novel category alignment problem, this work provides an effective and practical solution for FGGCD.





\printcredits


\section*{Declaration of Competing Interest}
The authors declare that they have no known competing financial interests or personal relationships that could have appeared to influence the work reported in this paper.

\section*{Acknowledgements}
This research was supported by Shenzhen Fundamental Research Program (JCYJ20230807094104009)

\section*{Data availability}
The data used in this research are publicly available graph datasets.

\bibliographystyle{cas-model2-names}

\bibliography{cas-refs}



\end{document}